%% file: main.tex
\documentclass[10pt,twocolumn,letterpaper]{article}

\usepackage{cvpr}
\usepackage{times}
\usepackage{epsfig}
\usepackage{graphicx}
\usepackage{amsmath}
\usepackage{amssymb}

\usepackage{algorithm}
\usepackage{algorithmic}
\usepackage{marvosym}
\usepackage[numbers]{natbib}
\usepackage{multirow}
\usepackage{graphicx}
\usepackage{color}
\usepackage[it,small]{caption}
\usepackage{subcaption}
\usepackage{dsfont}
\usepackage{enumitem}
\usepackage{authblk}
\newcommand{\ve}[1]{\mathbf{#1}}
\newcommand{\mcal}[1]{\mathcal{#1}}
\newcommand{\rc}[1]{\textcolor{red}{#1}}

\usepackage[pagebackref=true,breaklinks=true,letterpaper=true,colorlinks,bookmarks=false]{hyperref}

\def\cvprPaperID{1033} % *** Enter the CVPR Paper ID here

\cvprfinalcopy

\input{space_saver}
\begin{document}

\title{ Structural-RNN: Deep Learning on Spatio-Temporal Graphs}
\author[1,2]{Ashesh Jain}
\author[2]{Amir R. Zamir}
\author[2]{Silvio Savarese}
\author[3]{Ashutosh Saxena}
\affil[ ]{Cornell University$^1$, Stanford University$^2$, Brain Of Things Inc.$^3$}
\affil[ ]{{ashesh@cs.cornell.edu, \{zamir,ssilvio,asaxena\}@cs.stanford.edu}}

\iffalse
\title{ Structural-RNN: Deep Learning on Spatio-Temporal Graphs}
\author{First Author\\
Institution1\\
Institution1 address\\
{\tt\small firstauthor@i1.org}
\and
Second Author\\
Institution2\\
First line of institution2 address\\
{\tt\small secondauthor@i2.org}
}
\fi

\maketitle

\input{abstract}

%\input{intro_new}

\input{intro}

\input{relatedwork}

\input{model}

\input{experiment}

\input{conclusion}

{\small
\bibliographystyle{ieee}
\bibliography{shortstrings,references}
}

\end{document}

\title{ Supplementary for\\  Structural-RNN: Deep Learning on Spatio-Temporal Graphs}
\author[1,2]{Ashesh Jain}
\author[2]{Amir R. Zamir}
\author[2]{Silvio Savarese}
\author[3]{Ashutosh Saxena}
\affil[ ]{Cornell University$^1$, Stanford University$^2$, Brain Of Things Inc.$^3$}
\affil[ ]{{ashesh@cs.cornell.edu, \{zamir,ssilvio,asaxena\}@cs.stanford.edu}}

\iffalse
\title{ Supplementary for\\ Structural-RNN: Deep Learning on Spatio-Temporal Graphs}
\author{First Author\\
Institution1\\
Institution1 address\\
{\tt\small firstauthor@i1.org}
\and
Second Author\\
Institution2\\
First line of institution2 address\\
{\tt\small secondauthor@i2.org}
}
\fi

\maketitle

\setcounter{section}{3}
\section{Experiments}
\subsection{Human motion modeling and forecasting}
\noindent \textbf{User study.}  We randomly sampled three seed motions from each of the four activities (walking, eating, smoking, and discussion), giving a total of 12 seed motions. We forecasted human motion from the seeds using S-RNN, LSTM-3LR and ERD, resulting in total of 36 forecasted motions -- equally divided across algorithms and activities. We asked five users to rate the forecasted motions on a Likert scale of 1 -- 3, where a score of 1 is bad, 2 is neutral, and 3 is good. The users were instructed to rate  based on how human like the forecasted motion appeared. In order to calibrate, the users were first shown many examples of ground truth motion capture videos.

Figure~\ref{fig:user_study} shows the number of examples that obtained bad, neutral, and good scores for each algorithm. Majority of the motions generated by S-RNN were of high-quality and resembled human like motion. On the other hand, LSTM-3LR generated reasonable motions most of the times, however they were not as good as the ground truth. Finally, the motions forecasted by ERD were not human like for most of the aperiodic activities (eating, smoking, and discussion). On the walking activity, all algorithms were competitive and users mostly gave a score of 3 (good). Hence, through the user study we validate that S-RNN generates most realistic human motions majority of the times. Look at the supplementary video for more details. 

\noindent \textbf{Training S-RNN for motion forecasting.} We closely follow the training procedure by Fragkiadaki et al.~\citep{Fragkiadaki15}. We cross-validate over the hyperparameters on the validation set and set them to the following values:
\begin{itemize}
\item Back propagation through 100 time steps.
\item Mini-batch size of 100 sequences.
\item We use SGD and start with the step-size of $10^{-3}$. We decay the step-size by 0.1 when the training error plateaus. 
\item We clip the L2-norm of gradient to 25.0, and clip each dimension to [-5.0, 5.0]
\item We gradually add Gaussian noise to the training data following the schedule: at iterations \{250, 500, 1000, 1300, 2000, 2500, 3300\} we add noise with standard deviation \{0.01, 0.05, 0.1, 0.2, 0.3, 0.5, 0.7\}. As also noted by Fragkiadaki et al.~\citep{Fragkiadaki15}, adding noise is very important during training in order to forecast motions that lie on the manifold of human-like motions. 
\end{itemize}

\begin{figure}[t]
\centering
\includegraphics[width=0.8\linewidth]{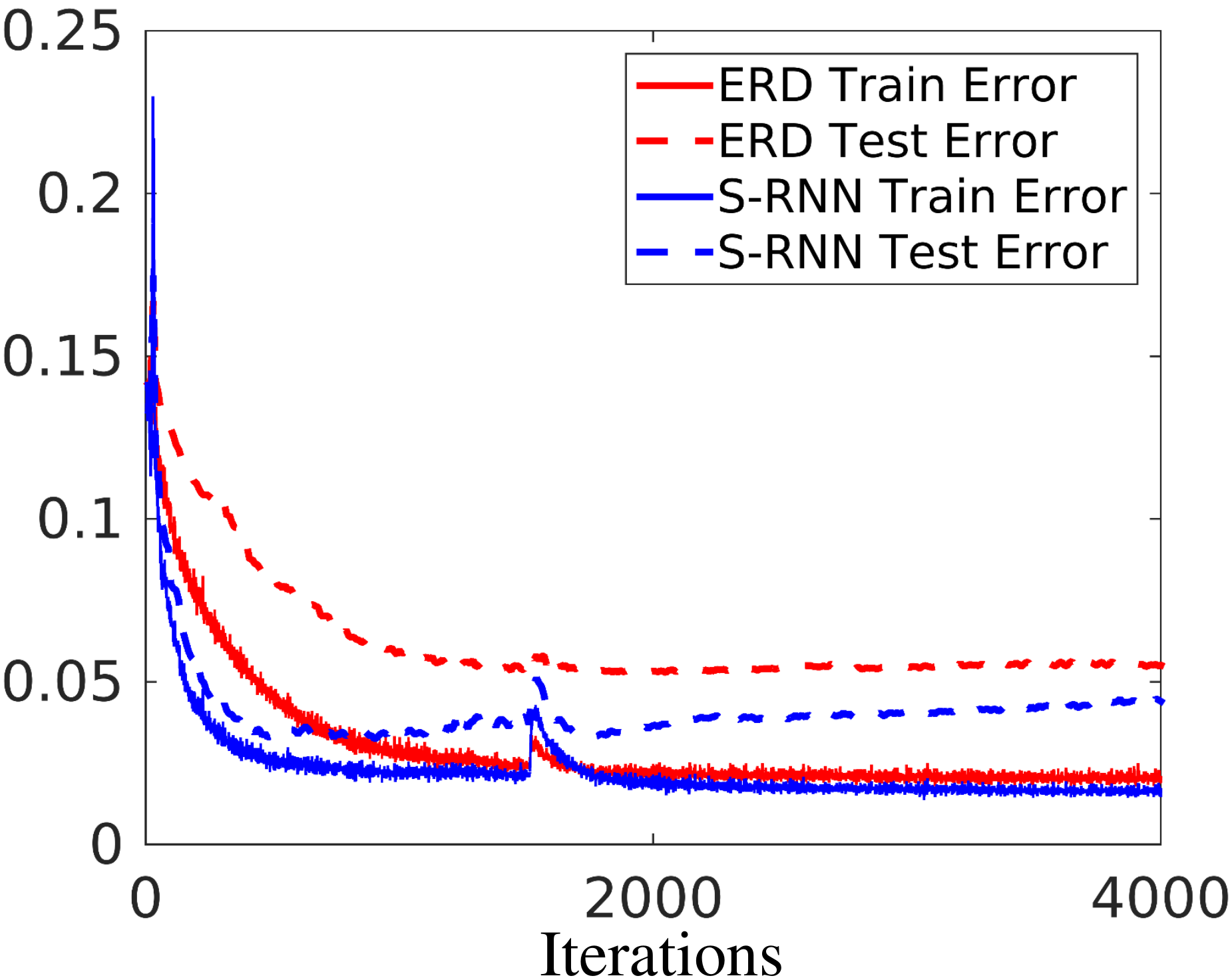}
\caption{ \textbf{Train and test error}. S-RNN generalizes better than ERD with a smaller test error. }
%Structural nature of our architecture allows us to exchange nodeRNNs between S-RNN architectures trained on different motion styles. 
\label{fig:loss}
\end{figure}

Figure~\ref{fig:loss} examines the test and train error with iterations. Both S-RNN and ERD converge to similar training error, however S-RNN generalizes better with a smaller test error for next step prediction. The number of parameters in S-RNN are marginally more than ERD. S-RNN have more LSTMs than ERD, but each LSTM in S-RNN is half the size of the LSTMs used in ERD. For ERD we used the best set of parameters described in~\citep{Fragkiadaki15}. There, the authors cross-validated over model parameters. In the plot, the  jump in error around iteration 1500 corresponds to the decay in step size. Due to addition of noise the test error of S-RNN exhibits a small positive slope, but it always stays below ERD.

\begin{figure}[t]
\centering
%\vspace{\sectionReduceTop}
\includegraphics[width=.8\linewidth]{../Figs/user_study.pdf}
\caption{\textbf{User study} with five users. Each user was shown 36 forecasted motions equally divided across four activities (walking, eating, smoking, discussion) and three algorithms (S-RNN, ERD, LSTM-3LR). The plot shows the number of bad, neutral, and good motions forecasted by each algorithm.}
\label{fig:user_study}
\end{figure}

\setcounter{subsection}{3}

\begin{table*}[t]
\centering
\caption{\footnotesize{\textbf{Maneuver Anticipation on 1100 miles of real-world driving data}. S-RNN is derived from the st-graph shown in Figure~\ref{fig:stgraphexp}\rc{c}. Jain et al.~\cite{Jain15} use the same st-graph but models it in a probabilistic frame with AIO-HMM. The table shows average \textit{precision}, \textit{recall} and \textit{time-to-maneuver}. Time-to-maneuver is the interval between the time of algorithm's prediction and the start of the maneuver. Algorithms are compared on the features from~\cite{Jain15}. }}
\vspace{1\sectionReduceBot}
\resizebox{0.9\linewidth}{!}{
\centering
\begin{tabular}{cr|ccc|ccc|ccc}
%\hline
&  &\multicolumn{3}{c}{Turns}&\multicolumn{3}{|c}{Lane change}&\multicolumn{3}{|c}{All maneuvers}\\
\cline{1-11}
\multicolumn{2}{c|}{\multirow{2}{*}{Method}} & \multirow{2}{*}{$Pr$ (\%)}  & \multirow{2}{*}{$Re$ (\%)} & Time-to-  & \multirow{2}{*}{$Pr$ (\%)} & \multirow{2}{*}{$Re$ (\%)} & Time-to-  & \multirow{2}{*}{$Pr$ (\%)} & \multirow{2}{*}{$Re$ (\%)}  & Time-to- \\ 
& & & &  maneuver (s) &  & &  maneuver (s) &  & & maneuver (s)\\\hline
&SVM 		&	64.7 &	47.2 &	2.40 	&	73.7 &	57.8 &	2.40		&	43.7 &	37.7 & 1.20\\
&AIO-HMM	~\cite{Jain15}	 		&	{80.8}	&		75.2	&	4.16 &	{83.8}	&	79.2	&	3.80 		&	{77.4}	&	{71.2}	&	3.53 \\
 & S-RNN w/o edgeRNN  & 75.2 & 75.3  & 3.68 & 85.4  & \textbf{86.0} & 3.53 & 78.0 & 71.1 & 3.15 \\
&(Ours) S-RNN  & \textbf{81.2}  & \textbf{78.6} & 3.94 & \textbf{92.7} & 84.4  & 3.46 & \textbf{82.2}  & \textbf{75.9} & 3.75 \\\hline
%\textit{Methods}&S-RNN with EL & 88.2 $\pm$ 1.4 & \textbf{86.0} $\pm$ 0.7 & 3.42 & \textbf{83.8} $\pm$ 2.1 & \textbf{79.9} $\pm$ 3.5 & 3.78 & \textbf{84.5} $\pm$ 1.0 & \textbf{77.1} $\pm$ 1.3 & 3.58\\
%\hline
\end{tabular}
}
\vspace{2\sectionReduceTop}
\label{tab:maneuver}
\end{table*}

\subsection{Driving maneuver anticipation}

\begin{figure}[t]
\centering
%\vspace{\sectionReduceTop}
\includegraphics[width=.9\linewidth]{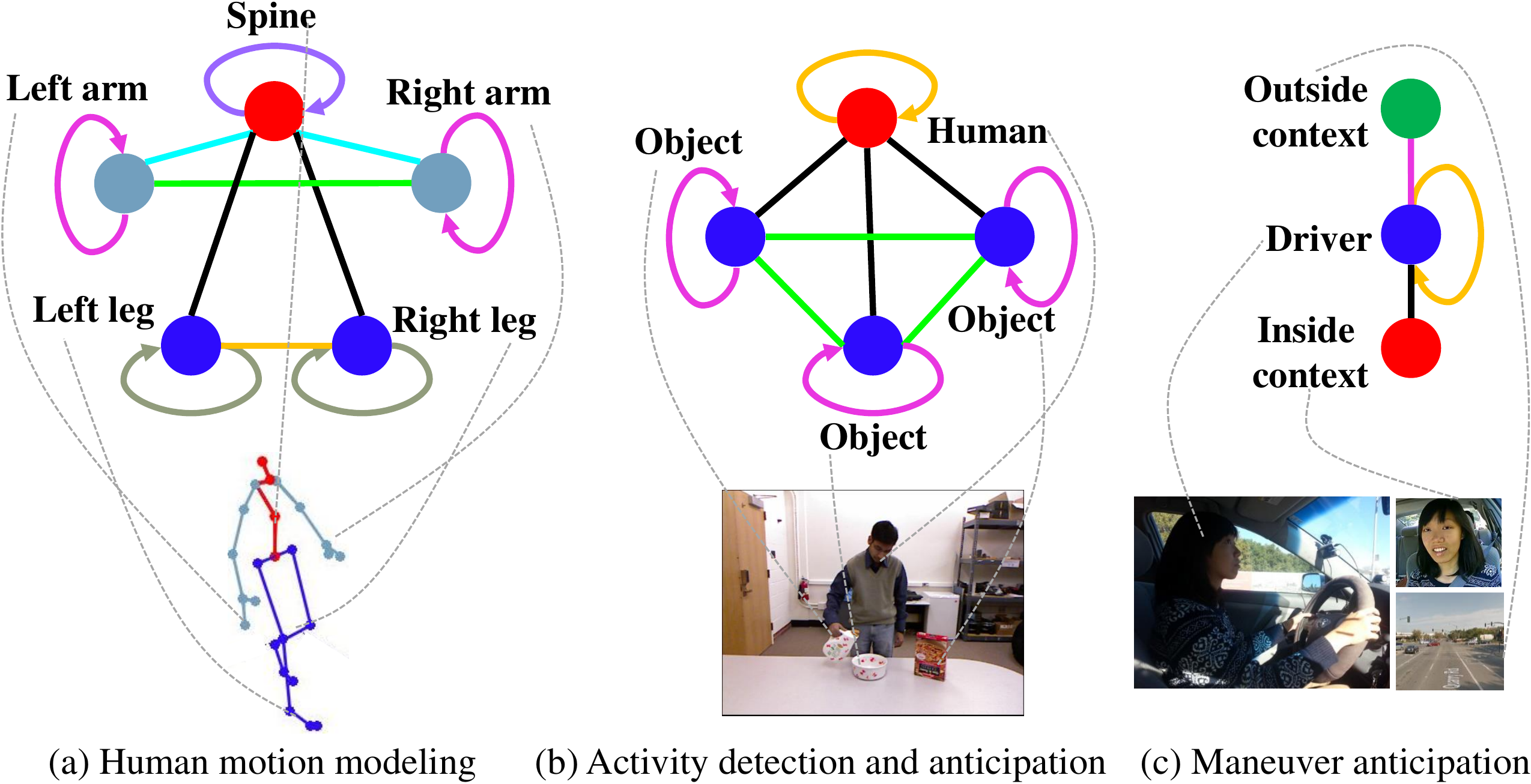}
\caption{\footnotesize{\textbf{Diverse spatio-temporal tasks}. We apply S-RNN to the following three diverse spatio-temporal problems. (View in color)}}
\label{fig:stgraphexp}
\end{figure}

We now present S-RNN for another application which involves anticipating maneuvers several seconds before they happen. For example, anticipating a future lane change maneuver several seconds before the wheel touches the lane markings. This problem requires spatial and temporal reasoning of the driver, and the sensory observations from inside and outside of the car. Jain et al.~\cite{Jain15} represent this problem with the st-graph shown in Figure~\ref{fig:stgraphexp}\rc{c}. They model the st-graph as a probabilistic Bayesian network called AIO-HMM. The st-graph represents the interactions between the observations outside the vehicle (eg. the road features), the driver's maneuvers, and the observations inside the vehicle (eg. the driver's facial features). We model the same st-graph with S-RNN architecture using the node and edge features from Jain et al.~\cite{Jain15}.

The nodeRNN models the driver, and the two edgeRNNs model the interactions between the driver and the observations inside the vehicle, and the observations outside the vehicle. The driver node is labeled with the future maneuver and, the observation nodes do not carry any label. The output of the driver nodeRNN is softmax probabilities of the following five maneuvers: \{\textit{Left lane change, right lane change, left turn, right turn, straight driving}\}.  Our nodeRNN architecture is RNN(64)-softmax(5), and edgeRNN is LSTM(64).

We train S-RNN on the features provided by Jain et al.~\cite{Jain15} on their 1100 miles of natural driving data set. The algorithms are evaluated on their precision and recall in anticipating maneuvers  under the following three prediction
settings: (i) Lane change: algorithms only anticipate lane changes. This setting is relevant for freeway driving; (ii) Turns: algorithms only anticipate turns; and (iii) All maneuvers: algorithms anticipate all five maneuvers. Table~\ref{tab:maneuver} shows the performance of different algorithms on this task.  S-RNN performs better than the previous state-of-the-art AIO-HMM~\cite{Jain15} in every setting.  It improves the precision  by 5\% and recall by 4\% with predicting all five maneuvers. Both AIO-HMM and S-RNN model the same st-graph but using different techniques. The table also shows that the performance decreases if we remove edgeRNNs and simply feed the concatenation of edge features into the nodeRNN. This emphasizes importance of the edgeRNNs,  and the need for separately modeling different kinds of edge interactions.

\iffalse
\begin{table*}[t]
\centering
\caption{\footnotesize{\textbf{Detection and anticipation results on
CAD-120~\cite{Koppula13b}}. S-RNN architecture derived from st-graph in
Figure~\ref{fig:stgraphexp}b outperforms Koppula et al.~\cite{Koppula13b} which
models the st-graph in a probabilistic framework. S-RNN in multi-task setting
(joint detection and anticipation) further improves the performance. \todo{This
will go to supplementary.}}}
%\resizebox{1\linewidth}{!}{
	\centering
	\begin{tabular}{r|cccc|cccc}
	Methods&\multicolumn{4}{c|}{Detection}&
	\multicolumn{4}{c}{Anticipation}\\\hline
	 & \multicolumn{2}{c}{Sub-activity} & \multicolumn{2}{c|}{Object
	 affordance} & \multicolumn{2}{c}{Sub-activity} &
	 \multicolumn{2}{c}{Object affordance} \\
	 & Micro P/R  & F1-score & Micro P/R  & F1-score & Micro P/R  & F1-score
	 & Micro P/R  & F1-score\\\hline 
	 Koppula et
	 al.~\cite{Koppula13,Koppula13b}&86.0&80.4&91.8&81.5&47.7&37.9&66.1&36.7\\
	 S-RNN w/o edgeRNN &83.5&82.2&90.1&82.1&70.5&64.8&77.0&72.4\\
	 S-RNN &85.6&83.2&92.4&88.7&69.1&62.3&81.9&80.7\\
	 S-RNN (multi-task) &85.7&82.4&92.8&91.1&72.3&65.7&81.5&80.9\\
	 %&activity (\%)&Affordance (\%)&activity (\%)&Affordance (\%)\\\hline
	 %Koppula et
	 %al.~\cite{Koppula13,Koppula13b}&\textbf{86.0}&91.8&47.7&66.1\\
	 %\todo{w/o edgeRNN} &&&&\\
	 %S-RNN &85.5&92.4&68.9&\textbf{81.8}\\
	 %S-RNN (multi-task)&85.7&\textbf{92.8}&\textbf{72.3}&81.5\\\hline
	 \end{tabular}
	 %}
	 \label{tab:cad120-supp}

	 \end{table*}

\fi

{\small
\bibliographystyle{ieee}
\bibliography{../COMTools/shortstrings,../COMTools/references}
}

%% file: space_saver.tex
 \belowdisplayskip \abovedisplayskip
\renewcommand{\baselinestretch}{0.97}

 \newlength\savedwidth

\selectfont

\newlength{\sectionReduceTop}
\newlength{\sectionReduceBot}
\newlength{\subsectionReduceTop}
\newlength{\subsectionReduceBot}
\newlength{\abstractReduceTop}
\newlength{\abstractReduceBot}
\newlength{\captionReduceTop}
\newlength{\captionReduceBot}
\newlength{\subsubsectionReduceTop}
\newlength{\subsubsectionReduceBot}

\newlength{\horSkip}
\newlength{\verSkip}

\newlength{\figureHeight}

%% file: abstract.tex
\begin{abstract}
Deep Recurrent Neural Network architectures, though remarkably capable at modeling sequences, lack an intuitive high-level spatio-temporal structure. That is while many problems in computer vision inherently have an underlying high-level structure and can benefit from it. Spatio-temporal graphs are a popular  tool for imposing such high-level intuitions in the formulation of real world problems. In this paper, we propose an approach for combining the power of high-level spatio-temporal graphs and sequence learning success of Recurrent Neural Networks~(RNNs). We develop a scalable method for casting an arbitrary spatio-temporal graph as a rich RNN mixture that is feedforward, fully differentiable, and jointly trainable. The proposed method is generic and principled as it can be used for transforming any spatio-temporal graph through employing a certain set of well defined steps. The evaluations of the proposed approach on a diverse set of problems, ranging from modeling human motion to object interactions, shows improvement over the state-of-the-art with a large margin. We expect this method to empower  new approaches to problem formulation through high-level spatio-temporal graphs and Recurrent Neural Networks.
\end{abstract}
%This method is expected to empower a new approach to problem formulation through high-level spatio-temporal graphs and Recurrent Neural Networks and be of broad interest to the community.

%% file: intro.tex
%\noindent\textbf{Links:} \href{https://cs.stanford.edu/people/ashesh/srnn}{\Mundus Web}
\noindent\textbf{Links:} \href{http://asheshjain.org/srnn}{\Mundus Web}
\\~
\vspace{2\sectionReduceTop}
\section{Introduction}
%\vspace{1\sectionReduceBot}
\begin{figure}[t]
\centering
\includegraphics[width=.9\linewidth]{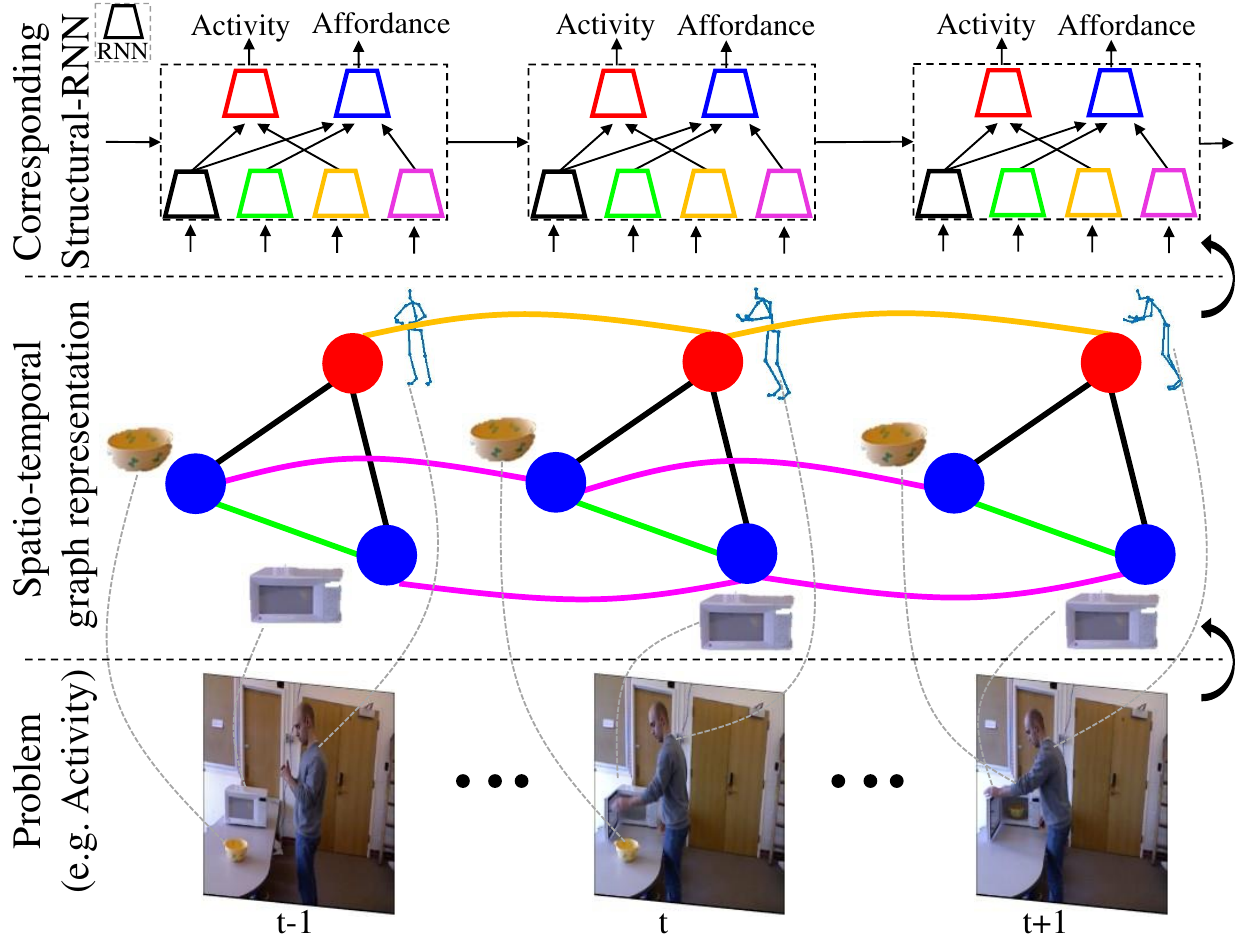}
\vspace{1.0\sectionReduceTop}
\caption{\footnotesize{\textbf{From st-graph to S-RNN for an example problem.} \textbf{(Bottom)} Shows an example activity (human microwaving food). Modeling such problems requires both spatial and temporal reasoning. \textbf{(Middle)} St-graph capturing spatial and temporal interactions between the human and the objects. \textbf{(Top)} Schematic representation of our structural-RNN architecture automatically derived from st-graph. It captures the structure and interactions of st-graph in a rich yet scalable manner.}}
\vspace{1\sectionReduceBot}
\label{fig:pullfig}
\end{figure}

\iffalse
The world we live in is inherently structured. It is comprised of components that interact with each other in space and time. Utilizing these spatio-temporal structures when formulating problems allows domain experts to inject their high-level knowledge into learning frameworks. This has been the incentive for many efforts in computer vision and machine learning, such as Logic Networks~\cite{Richardson06}, Graphical Models~\cite{Koller09}, and Structured SVMs~\cite{Joachims09}. Such structures are of particular interest to computer vision and robotics communities because interactions between humans and environment in real world are inherently
spatio-temporal in nature.
\fi

The world we live in is inherently structured. It is comprised of components that interact with each other in space and time, leading to a spatio-temporal composition. Utilizing such structures in problem formulation allows domain-experts to inject their high-level knowledge in learning frameworks. This has been the incentive for many efforts in computer
vision and machine learning, such as Logic Networks~\cite{Richardson06}, Graphical Models~\cite{Koller09}, and Structured SVMs~\cite{Joachims09}. Structures that span over both space and time ({spatio-temporal}) are of particular interest to computer vision and robotics communities. Primarily, interactions between humans and environment in real world are inherently {spatio-temporal} in nature. 
For example, during a cooking activity, humans interact with multiple objects both in space and through time. Similarly, parts of human body (arms, legs, etc.) have individual functions but work with each other in concert to generate physically sensible motions.   
Hence, bringing high-level {spatio-temporal} structures and rich sequence modeling capabilities together is of particular importance for many applications. 
%In this work, we propose an approach for modeling high-level spatio-temporal problems through rich mixtures of Recurrent Neural Networks (RNNs).

The notable success of RNNs has proven their capability on many end-to-end learning tasks~\cite{Graves14,Fragkiadaki15,Donahue15,Zheng15}. However, they lack a high-level and intuitive spatio-temporal structure though they have been shown to be successful at modeling long sequences~\citep{Srivastava15,Mikolov14,Sutskever14}. Therefore, augmenting a high-level structure with learning capability of RNNs leads to a powerful tool that has the best of both worlds.
Spatio-temporal graphs (st-graphs) are a popular~\cite{Li08,Lezama11,Brendel11,Douillard11,Koppula13c,Zhang14b,Jain15} general tool for representing such high-level spatio-temporal structures. The nodes of the graph typically represent the problem components, and the edges capture their spatio-temporal interactions. To achieve the above goal, we develop a generic tool for transforming an arbitrary st-graph into a feedforward mixture of RNNs, named structural-RNN (S-RNN). Figure~\ref{fig:pullfig} schematically illustrates this process, where a sample spatio-temporal problem is shown at the bottom, the corresponding st-graph representation is shown in the middle, and our RNN mixture counterpart of the st-graph is shown at the top.

In high-level steps, given an arbitrary st-graph, we first roll it out in time and decompose it into a set of contributing factor components. The factors identify the independent components that collectively determine one decision and are derived from both edges and nodes of the st-graph. We then semantically group the factor components and represent each group using one RNN, which results in the desired RNN mixture. The main challenges of this transformation problem are: 1)  making the RNN mixture as rich as possible to enable learning complex functions, yet 2) keeping the RNN mixture scalable with respect to size of the input st-graph. In order to make the resulting RNN mixture rich, we liberally represent each spatio-temporal factor (including node factors, temporal edge factors, and spatio-temporal edge factors) using one RNN. On the other hand, to keep the overall mixture scalable but not lose the essential learning capacity, we utilize ``factor sharing'' (aka clique templates~\cite{Taskar02,Mccallum09,Sutton11}) and allow the factors with similar semantic functions to share an RNN.  This results in a rich and scalable feedforward mixture of RNNs that is equivalent to the provided st-graph in terms of input, output, and spatio-temporal relationships. The mixture is also fully differentiable, and therefore, can be trained jointly as one entity.

The proposed method is principled and generic as the transformation is based on a set of well defined steps and it is applicable to any problem that can be formulated as st-graphs (as defined in Section~\ref{sec:srnn}). Several previous works have attempted solving specific problems using a collection of RNNs~\cite{Srivastava15,Du15,Venugopalan14,Donahue15,Byeon15}, but they are almost unanimously task-specific. They also do not utilize mechanisms similar to factorization or factor sharing in devising their architecture to ensure richness and scalability.

%Our approach allows domain-experts to cast their problems as st-graphs and learn deep recurrent structures over them.
 S-RNN is also modular, as it is enjoying an underlying high-level structure. This enables easy high-level manipulations which are basically not possible in unstructured (plain-vanilla) RNNs (e.g., we will experimentally show forming a feasible hybrid human motion by mixing parts of different motion styles - Sec~\ref{sec:visualize} ).
We evaluate the proposed approach on a diverse set of spatio-temporal problems (human pose modeling and forecasting, human-object interaction, and driver decision making), and show significant improvements over the state of the art on each problem. We also study complexity and convergence properties of S-RNN and provide further experimental insights by visualizing its memory cells that reveals some cells interestingly represent certain semantic operations. The code of the entire framework that accepts a st-graph as the input and yields the output RNN mixture is available at the {\hbox{\texttt{\url{http://asheshjain.org/srnn}}}}.

The contribution of this paper are: 1) a generic method for casting an arbitrary st-graph as a rich, scalable, and jointly trainable RNN mixture, 2) in defence of structured approaches, we show S-RNN significantly outperforms its unstructured (plain-vanilla) RNN counterparts, 3) in defence of RNNs, we show on several diverse spatio-temporal problems that modeling structure with S-RNN outperforms the non-deep learning based structured counterparts.%, such as probabilistic approaches.

%% file: relatedwork.tex
%\vspace{1\sectionReduceTop}
\section{Related Work}
%\vspace{1\sectionReduceBot}
We give a  categorized overview of the related literature.
%space removal
In general, three main characteristics differentiate our work from existing techniques: being generic and not restricted to a specific problem, providing a principled method for transforming a st-graph into a scalable rich feedforward RNN mixture, and being jointly trainable. 

\begin{figure*}[t]
\centering
\includegraphics[width=\linewidth]{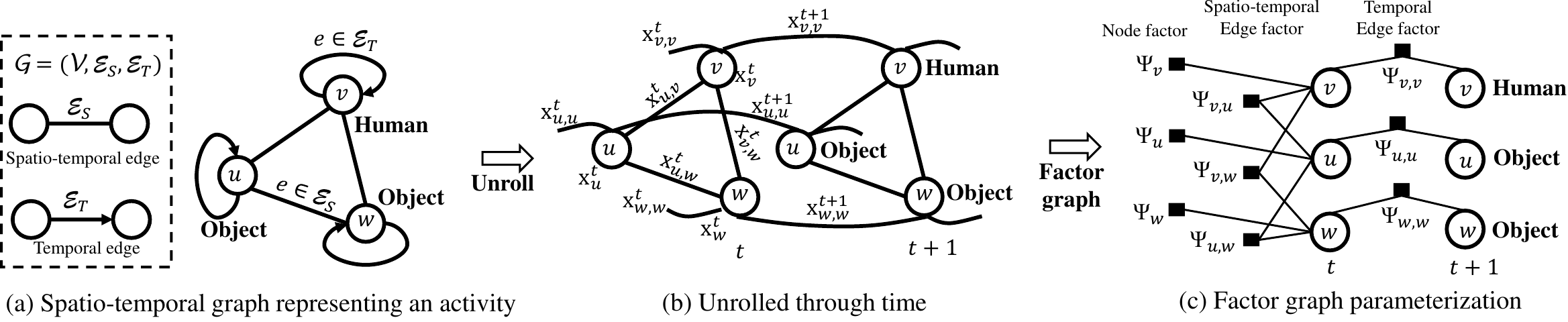}
\vspace{3\sectionReduceTop}
\caption{\footnotesize{\textbf{An example spatio-temporal graph (st-graph) of a human activity.} (a) st-graph capturing human-object interaction. (b) Unrolling the st-graph through edges $\mcal{E}_T$. The nodes and edges are labelled with the feature vectors associated with them. (c) Our factor graph parameterization of the st-graph. Each node and edge in the st-graph has a corresponding factor.}}
\label{fig:stgraph2}
\vspace{2\sectionReduceBot}
\end{figure*}
%Later we will present our structural-RNN architecture for modeling such spatio-temporal reasoning. 

\textit{Spatio-temporal problems.} 
%Our work is motivated with the need of richer learning architectures for problems represented as spatio-temporal graphs (``st-graphs''). 
Problems that require spatial and temporal reasoning are very common in robotics and computer vision. Examples include human activity recognition and segmentation from videos~\cite{Sun13,Shi11,Wang13b,Vail07,Chen09,Jain15b,Lezama11,Laptev08}, context-rich human-object interactions~\cite{Li08,Koppula15,Koppula13b,Gupta09,Koppula14}, modeling human motion~\cite{Fragkiadaki15,Taylor06,Taylor10} etc. Spatio-temporal reasoning also finds application in assistive robots, driver understanding, and object recognition~\cite{Zhang14b,Jain15,Pieropan14,Jain13b,Douillard11}. In fact most of our daily activities are spatio-temporal in nature. With growing interests in rich interactions and robotics, this form of reasoning will become even more important. We evaluate our generic method on three context-rich spatio-temporal problems: (i) Human motion modeling~\cite{Fragkiadaki15}; (ii) Human-object interaction understanding~\cite{Koppula15}; and (iii) Driver maneuver anticipation~\cite{Jain15}.

%Our approach is generic and we demonstrate it on the 
\textit{Mixtures of deep architectures.} Several previous works build multiple networks and wire them together in order to capture some complex structure (or interactions) in the problem with promising results on applications such as activity detection, scene labeling, image captioning, and object detection~\cite{Du15,Byeon15,Chen15,Girshick15,Srivastava15,Venugopalan14}. 
%They motivate the use of architectures with expressive structures. 
However, such architectures are mostly hand designed for specific problems, though they demonstrate the benefit in using a modular deep architecture. 
%For a new application at hand, practitioners have to redesign (and implement) the architecture from scratch. 
Recursive Neural Networks~\cite{Goller96} are, on the other hand, generic feedforward architectures, but for problems with recursive structure such as parsing of natural sentences and scenes~\cite{Socher11}. Our work is a remedy for problems expressed as spatio-temporal graphs. For a new spatio-temporal problem in hand, all a practitioner needs to do is to express their intuition about the problem as an st-graph.
	
\textit{Deep learning with graphical models.} Many works have addressed deep networks with graphical models for structured prediction tasks. Bengio et al.~\cite{Bengio94} combined CNNs with HMM for hand writing recognition. Tompson et al.~\cite{Tompson14} jointly train CNN and MRF for human pose estimation. Chen et al.~\cite{Chen14b} use a similar approach for image classification with general MRF. Recently several works have addressed end-to-end image segmentation with fully connected CRF~\cite{Zheng15,Liu15,Schwing15,Lin15}. Several works follow a two-stage approach and decouple the deep network from CRF. They have been applied to multiple problems including image segmentation, pose estimation, document processing~\cite{Zhang14,Chen14,Li15,Bottou97} etc. All of these works advocate and well demonstrate the benefit in exploiting the structure in the problem together with rich deep architectures. However, they largely do not address spatio-temporal problems and the proposed architectures are task-specific. 
%Our approach is not tied to a single problem, but it is applicable to any problem expressed over st-graph. We focus on context-rich spatio-temporal applications and  broaden the spectrum of problems addressed with structure of deep architectures.

%All these works advocate for exploiting structure in the problem together with end-to-end trainable rich architectures. These works does not address problems involving rich spatio-temporal structures, and the architectures proposed were largely problem specific. We propose a generic architecture for problems that can be expressed as spatio-temporal graph. We focus on context-rich spatio-temporal applications and  broaden the spectrum of problems addressed with structure of deep architectures.  

\textit{Conditional Random Fields (CRF)} model dependencies between the outputs by learning a joint distribution over them. They have been applied to many applications~\cite{Krahenbuhl12,Felzenszwalb08,Quattoni04} including st-graphs which are commonly modeled as spatio-temporal CRF~\cite{Li08,Koppula15,Zhang14b,Douillard11}. In our approach, we adopt st-graphs as a general graph representation and embody it using an RNN mixture architecture. Unlike CRF, our approach is not probabilistic and is not meant to model the joint distribution over the outputs. S-RNN instead learns the dependencies between the outputs via structural sharing of RNNs between the outputs. %(\todo{Amir, move this to your favourite location.})

%% file: model.tex
\begin{figure*}[t]
\centering
\includegraphics[width=\linewidth]{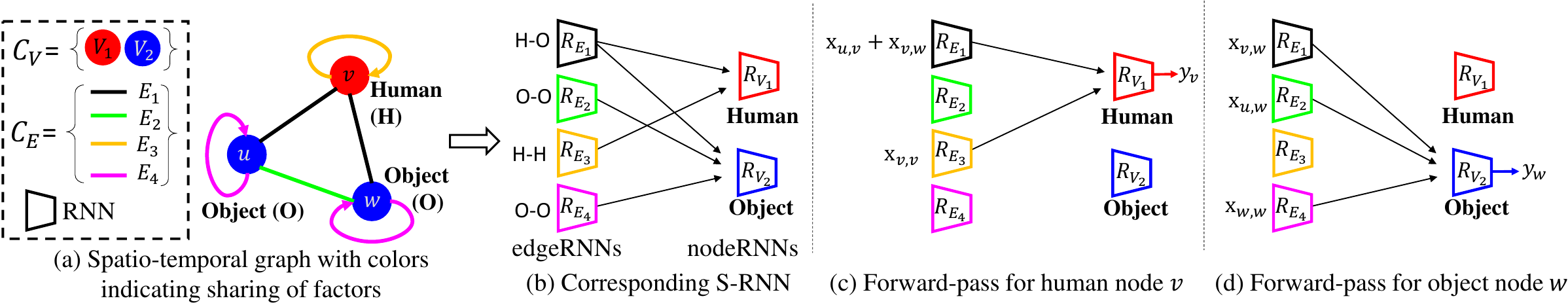}
\vspace{3\sectionReduceTop}
\caption{\footnotesize{\textbf{An example of st-graph to S-RNN.} \textbf{(a)} The st-graph from Figure~\ref{fig:stgraph2} is redrawn with colors to indicate sharing of nodes and edge factors. Nodes and edges with same color share factors. Overall there are six distinct factors: 2 node factors and 4 edge factors. \textbf{(b)} S-RNN architecture has one RNN for each factor. EdgeRNNs and nodeRNNs are connected to form a bipartite graph. Parameter sharing between the human and object nodes happen through edgeRNN $\ve{R}_{E_1}$. \textbf{(c)} The forward-pass for human node $v$ involve RNNs $\ve{R}_{E_1}$, $\ve{R}_{{E}_3}$ and $\ve{R}_{V_1}$. In Figure~\ref{fig:strnn} we show the detailed layout of this forward-pass. Input features into  $\ve{R}_{E_1}$ is sum of human-object edge features $\ve{x}_{u,v}+\ve{x}_{v,w}$. \textbf{(d)} The forward-pass for object node $w$ involve RNNs $\ve{R}_{E_1}$, $\ve{R}_{{E}_2}$, $\ve{R}_{{E}_4}$ and $\ve{R}_{V_2}$. In this forward-pass,  the edgeRNN $\ve{R}_{E_1}$ only processes the edge feature $\ve{x}_{v,w}$. (Best viewed in color)}} %Therefore, during training, the edgeRNN $\ve{R}_{E_1}$ processes different linear combinations of the human-object edge features depending on the node being forward-passed.}% In this way S-RNN also captures structure in the feature space.}
\vspace{1.5\sectionReduceBot}
	\label{fig:stgraph}
\end{figure*}

%\vspace{1\sectionReduceTop}
\section{Structural-RNN architectures}
%\vspace{1\sectionReduceBot}
\label{sec:srnn}
In this section we describe our approach for building structural-RNN (S-RNN) architectures.  %The nodes of the graph represent entities that we care about and edges represent interactions between them. 
We start with a \mbox{st-graph}, decompose it into a set of factor components, then represent each factor using a RNN. 
The RNNs are interconnected in a way that the resulting architecture captures the structure and interactions of the st-graph. %space removal

\vspace{1\subsectionReduceTop}
\subsection{Representation of spatio-temporal graphs}
\vspace{1\subsectionReduceTop}
\label{sec:stgraph}
Many applications that require spatial and temporal reasoning are modeled using st-graphs~\cite{Brendel11,Douillard11,Koppula15,Zhang14b,Jain15}.
%A spatio-temporal graph ``intuitively'' captures interaction between entities which are represented as nodes in the graph. 
We represent a st-graph with $\mathcal{G}=(\mathcal{V},\mathcal{E}_S,\mathcal{E}_T)$,  whose structure $(\mathcal{V},\mathcal{E}_S)$ unrolls over time through edges $\mathcal{E}_T$. Figure~\ref{fig:stgraph2}\rc{a} shows an example st-graph capturing human-object interactions during an activity.  The nodes $v \in \mcal{V}$ and edges $e \in \mcal{E}_S\cup \mcal{E}_T$ of the st-graph repeats over time. In particular, Figure~\ref{fig:stgraph2}\rc{b} shows the same st-graph unrolled through time. In the unrolled st-graph, the nodes at a given time step $t$ are connected with undirected \textit{spatio-temporal} edge $e=(u,v) \in \mcal{E}_S$, and the nodes at adjacent time steps (say the node $u$ at time $t$ and the node $v$ at time $t+1$)  are connected with  undirected \textit{temporal} edge \textit{iff} $(u,v) \in \mcal{E}_T$.\footnote{For simplicity, the example st-graph in Figure~\ref{fig:stgraph2}\rc{a} considers temporal edges of the form  $(v,v) \in \mcal{E}_T$.}

Given a st-graph and the feature vectors associated with the nodes $\ve{x}_v^t$ and edges $\ve{x}_e^t$, as shown in Figure~\ref{fig:stgraph2}\rc{b}, the goal is to predict the node labels (or real value vectors) $y_v^t$ at each time step $t$. For instance, in human-object interaction, the node features can represent the human and object poses, and edge features can their relative orientation; the node labels represent the human activity and object affordance. 
%We propose a generic feedforward neural network architecture for this problem. The challenge lies in capturing the node and edge interactions represented by the st-graph.  
Label $y_v^t$ is affected by both its node and its interactions with other nodes (edges), leading to an overall complex system.
Such interactions  are commonly parameterized with a factor graph that conveys how a (complicated) function over the st-graph factorizes into simpler functions~\cite{Kschischang01}.
We derive our S-RNN architecture from the factor graph representation of the st-graph. Our factor graph representation has a factor function $\Psi_v(y_v,\ve{x}_v)$ for each node and a pairwise factor $\Psi_e(y_{e(1)},y_{e(2)},\ve{x}_e)$ for each edge. Figure~\ref{fig:stgraph2}\rc{c} shows the factor graph corresponding to the st-graph in ~\ref{fig:stgraph2}\rc{a}. \footnote{Note that we adopted factor graph as a tool for capturing interactions and not modeling the overall function. Factor graphs are commonly used in probabilistic graphical models for factorizing joint probability distributions. We consider them for general st-graphs and do not establish relations to its probabilistic and function decomposition properties.} %However an intuition of graphical models can be helpful in coming up with a factor graph.}

\iffalse
The structure of our architecture captures the interactions in the st-graph. Such interactions  are commonly parameterized with a factor graph. It conveys how a (complicated) function over st-graph factorizes into simpler function components~\cite{Kschischang01}.  We derive our S-RNN from the factor graph representation of the st-graph. S-RNN use factor graph as a tool for capturing interactions and do not model the function represented by the factor graph.\footnote{Factor graphs are commonly used in graphical models for factorizing joint distributions. We consider factor graphs for general st-graphs. However an intuition of graphical models can be helpful in coming up with a factor graph.} Our factor graph representation has a factor (function) $\Psi_v(y_v,\ve{x}_v)$ for each node, and a pairwise factor  $\Psi_e(y_{e(1)},y_{e(2)},\ve{x}_e)$ for each edge. Figure~\ref{fig:stgraph2}\rc{c} shows the factor graph representation of the st-graph in Figure~\ref{fig:stgraph2}\rc{a}.
\fi

\textbf{Sharing factors between nodes.}
Each factor in the st-graph has parameters that needs to be learned. Instead of learning a distinct factor for each node, semantically similar nodes can optionally share factors. For example, all ``object nodes'' \{$u$,$w$\} in the st-graph can share the same node factor and parameters. This modeling choice allows enforcing parameter sharing between similar nodes.
%to prevent increasing their number intractably with the size of the st-graph. 
It further gives the flexibility to handle st-graphs with more nodes without increasing the number of parameters. For this purpose, we partition the nodes as $\mcal{C}_V=\{V_1,..,V_P\}$ where $V_p$ is a set of semantically similar nodes, and they all use the same node factor $\Psi_{V_p}$. In Figure~\ref{fig:stgraph}\rc{a} we re-draw the st-graph and assign same color to the nodes sharing node factors. 

Partitioning nodes on their semantic meanings leads to a natural semantic partition of the edges, $\mcal{C}_E=\{E_1,..,E_M\}$, where $E_m$ is a set of edges whose nodes form a semantic pair. Therefore, all edges in the set $E_m$  share the same edge factor $\Psi_{E_m}$. For example all ``human-object edges'' \{$(v,u), (v,w)$\} are modeled with the same edge factor.  Sharing factors based on semantic meaning makes the overall parametrization compact. In fact, sharing parameters is necessary to address applications where the number of nodes depends on the context. For example, in human-object interaction the number of object nodes vary with the environment. Therefore, without sharing parameters between the object nodes, the model cannot generalize to new environments with more objects. For modeling flexibility, the edge factors are not shared across the edges in $\mcal{E}_S$ and $\mcal{E}_T$. Hence, in Figure~\ref{fig:stgraph}\rc{a}, object-object $(w,w)\in\mcal{E}_T$ temporal edge is colored differently from object-object $(u,w) \in \mcal{E}_S$ spatio-temporal edge.%In all there are $P$ distinct node factors and $M$ distinct edge factors. It also allows for incorporating training examples with more nodes (eg. objects) and edges, as long as they belong to one of the semantic categories. 

In order to predict the label of node $v \in V_p$, we consider its node factor $\Psi_{V_p}$, and the edge factors connected to $v$ in the factor graph. We define a node factor and an edge factor as neighbors if they jointly affect the label of some node in the st-graph. More formally, the node factor $\Psi_{V_p}$ and edge factor $\Psi_{E_m}$ are \textit{neighbors}, if there exist a node $v \in {V_p}$ such that it connects to both $\Psi_{V_p}$ and $\Psi_{E_m}$ in the factor graph. We will use this definition in building S-RNN architecture such that it captures the interactions in the st-graph.

\iffalse
A node's label in the st-graph is affected by the factors the node is connected to in the factor-graph. We define {neighbor factors} to group the factors which jointly affect the node labels.  The node factor $\Psi_{V_p}$ and edge factor $\Psi_{E_m}$ are \textit{neighbors}, if there exist a node $v \in {V}$ such that it connects to both $\Psi_{V_p}$ and $\Psi_{E_m}$ in the factor graph. This definition will be instrumental in building the S-RNN architecture such that it captures the interactions in the st-graph.

\begin{align}
	\nonumber \Psi_{V_p} \;\&\; \Psi_{E_m}\; &\text{are neighbors} \Longleftrightarrow \\
	\label{eq:neighbor} &\exists v \in V_p, u \in \mcal{V}\; \text{s.t.}\; (u,v) \in E_m
\end{align}
\fi

\vspace{1\subsectionReduceTop}
\subsection{Structural-RNN from spatio-temporal graphs}
\vspace{1\subsectionReduceBot}
\label{subsec:srnn}
We derive our S-RNN architecture from the factor graph representation of the st-graph. The factors in the st-graph operate in a temporal manner, where at each time step the factors observe (node \& edge) features and perform some computation on those features. In S-RNN, we represent each factor with an RNN. We refer the RNNs obtained from the node factors as nodeRNNs and the RNNs obtained from the edge factors as edgeRNNs. The interactions represented by the st-graph are captured through connections between the nodeRNNs and the edgeRNNs. 

\iffalse
We now present our S-RNN architecture to represent st-graphs described earlier.  Our contribution lies in constructively building an RNN ensemble, such that the RNNs and the connections between them are semantically related to the underlying st-graph. %Our architecture consists of multiple RNNs which are interconnected in a manner to capture the structure/interactions of the st-graph. This allows us, and will enable practitioners to learn expressive deep architectures for problems that are represented as spatio-temporal graphs. 
We derive our S-RNN architecture from the factor graph representation of the st-graph. The factors in the st-graph operate in a temporal manner, where at each time step the factors observe (node \& edge) features and perform some computation on those features. In S-RNN we represent each factor with an RNN. We refer the RNNs obtained from node factors as nodeRNNs and the RNNs obtained from edge factors as edgeRNNs. The interactions represented by the st-graph are captured through connections between the nodeRNNs and the edgeRNNs. 
\fi

We denote the RNNs corresponding to the node factor $\Psi_{V_p}$ and the edge factor $\Psi_{E_m}$ as  $\ve{R}_{V_p}$ and $\ve{R}_{E_m}$ respectively. In order to obtain a feedforward network, we connect the edgeRNNs and nodeRNNs to form a bipartite graph $\mcal{G}_R = (\{\ve{R}_{E_m}\},\{\ve{R}_{V_p}\},\mcal{E}_R)$. In particular, the edgeRNN $\ve{R}_{E_m}$ is connected to the nodeRNN $\ve{R}_{V_p}$ \textit{iff} the factors $\Psi_{E_m}$  and $\Psi_{V_p}$ are \textit{neighbors} in the st-graph, i.e. they jointly affect the label of some node in the st-graph. To summarize, in Algorithm~\ref{alg:srnn} we show the steps for constructing S-RNN architecture. 
Figure~\ref{fig:stgraph}\rc{b} shows the S-RNN for the human activity represented in  Figure~\ref{fig:stgraph}\rc{a}. The nodeRNNs combine the outputs of the edgeRNNs they are connected to (i.e. its \textit{neighbors} in the factor graph), and predict the node labels. The predictions of nodeRNNs (eg. $\ve{R}_{V_1}$ and $\ve{R}_{V_2}$) interact through the edgeRNNs (eg. $\ve{R}_{E_1}$).  Each edgeRNN handles a specific semantic interaction between the nodes connected in the st-grap and models how the interactions evolve over time. In the next section, we explain the inputs, outputs, and the training procedure of S-RNN.
%It is through edgeRNNs (eg. $\ve{R}_{E_1}$), the prediction of nodeRNNs ($\ve{R}_{V_1}$, $\ve{R}_{V_2}$) interact.
%Each edgeRNN handles a specific kind of semantic interaction between the nodes connected in the st-graph. Each nodeRNN combines the outputs of the edgeRNNs it is connected to (i.e. its \textit{neighbors} in the factor graph) and predicts the node label.  The resulting architecture captures the interactions represented by the st-graph. In the following section we explain the inputs, outputs, and the training procedure of S-RNN.
\begin{algorithm}[t]\caption{From spatio-temporal graph to S-RNN}
	\begin{algorithmic}
		\STATE \textbf{Input} $\mcal{G} = (\mcal{V},\mcal{E}_S,\mcal{E}_T)$, $C_V=\{V_1,...,V_P\}$ %, $C_E=\{E_1,...,E_M\}$
		\STATE \textbf{Output} S-RNN graph $\mcal{G}_R = (\{\ve{R}_{E_m}\},\{\ve{R}_{V_p}\},\mcal{E}_R)$
		\STATE \text{1:} Semantically partition edges $C_E=\{E_1,...,E_M\}$
		\STATE \text{2:} Find factor components $\{\Psi_{V_p},\Psi_{E_m}\}$  of $\mcal{G}$
		\STATE \text{3:} \text{Represent each $\Psi_{V_p}$ with a nodeRNN $\ve{R}_{V_p}$}
		\STATE \text{4:} \text{Represent each $\Psi_{E_m}$ with an edgeRNN $\ve{R}_{E_m}$}
		\STATE \text{5:} \text{Connect $\{\ve{R}_{E_m}\}$ and $\{\ve{R}_{V_p}\}$ to form a bipartite graph.}
		\STATE $(\ve{R}_{E_m}, \ve{R}_{V_p}) \in \mcal{E}_R$\; \textit{iff}\; $\exists v \in V_p, u \in \mcal{V}\; \text{s.t.}\; (u,v) \in E_m$
		\STATE \textbf{Return} $\mcal{G}_R = (\{\ve{R}_{E_m}\},\{\ve{R}_{V_p}\},\mcal{E}_R)$
	\end{algorithmic}
	\label{alg:srnn}
\end{algorithm}
\vspace{1\subsectionReduceTop}
\subsection{Training structural-RNN architecture}
\vspace{1\subsectionReduceBot}
In order to train the S-RNN architecture, for each node of the st-graph the features associated with the node are fed into the architecture.  In the forward-pass for node $v \in V_p$, the input into edgeRNN $\ve{R}_{E_m}$ is the temporal sequence of edge features $\ve{x}_e^t$ on the edge $e \in E_m$, where edge $e$ is incident to node $v$ in the st-graph.   The nodeRNN $\ve{R}_{V_p}$ at each time step concatenates the node feature $\ve{x}_v^t$ and the outputs of edgeRNNs it is connected to, and predicts the node label. At the time of training, the errors in prediction are back-propagated through the nodeRNN and edgeRNNs involved during the forward-pass. That way, S-RNN non-linearly combines the node and edge features associated with the nodes in order to predict the node labels.
%The output of edgeRNNs are high-level representations of the edge features.

\begin{figure}[t]
	\centering
	\includegraphics[width=.9\linewidth]{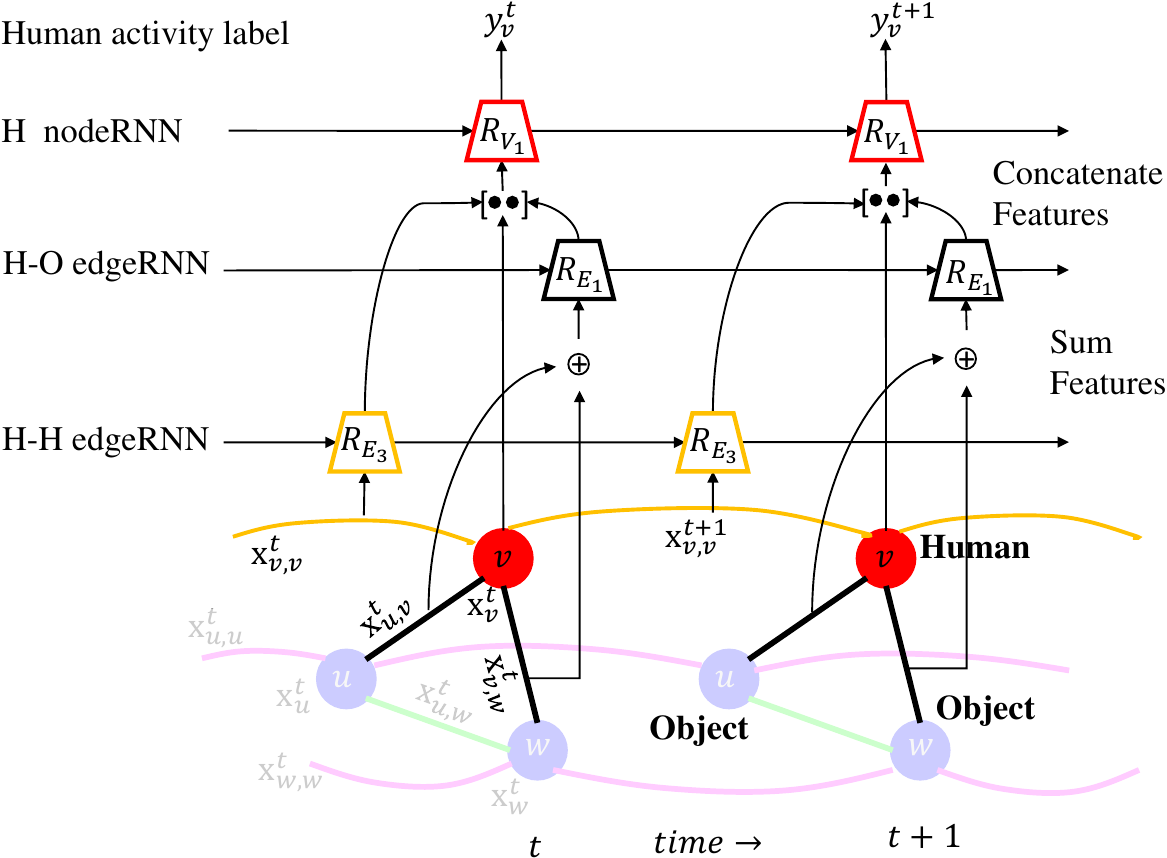}
	\vspace{1.5\sectionReduceTop}
	\caption{\footnotesize{\textbf{Forward-pass for human node $v$.} Shows the architecture layout corresponding to the Figure~\ref{fig:stgraph}\rc{c} on unrolled st-graph. (View in color)}}
	\label{fig:strnn}
	\vspace{2\sectionReduceBot}
\end{figure}

Figure~\ref{fig:stgraph}\rc{c} shows the forward-pass through S-RNN for the human node. Figure~\ref{fig:strnn} shows a detailed architecture layout of the same forward-pass. The forward-pass involves the edgeRNNs $\ve{R}_{E_1}$ (human-object edge) and $\ve{R}_{E_3}$ (human-human edge). Since the human node $v$ interacts with two object nodes \{$u$,$w$\},
we pass the summation of the two edge features as input to $\ve{R}_{E_1}$. The summation of features, as opposed to concatenation, is important to handle {variable number} of object nodes with a {fixed architecture}. Since the object count varies with environment, it is challenging to represent variable context with a fixed length feature vector. Empirically, adding features works better than mean pooling. We conjecture that addition retains the object count and the structure of the st-graph, while mean pooling averages out the number of edges.  The nodeRNN $\ve{R}_{V_1}$ concatenates the (human) node features with the outputs from edgeRNNs, and predicts the activity at each time step. 
% the input into the edgeRNN modeling this edge is the sum of two human-object edge feature vectors. The nodeRNN modeling the human node concatenates the node features with the output representations from edgeRNNs, and predicts the human activity. 

\textbf{Parameter sharing and structured feature space.} An important aspect of S-RNN is sharing of parameters across the node labels. Parameter sharing between node labels happen when an RNN is common in their forward-pass. For example in Figure~\ref{fig:stgraph}\rc{c}~and~\ref{fig:stgraph}\rc{d}, the edgeRNN $\ve{R}_{E_1}$ is common in the forward-pass for the human node and the object nodes. Furthermore, the parameters of $\ve{R}_{E_1}$ gets updated through back-propagated gradients from both the object and human nodeRNNs. In this way, $\ve{R}_{E_1}$ affects both the human and object node labels. 

Since the human node is connected to multiple object nodes, the input into edgeRNN $\ve{R}_{E_1}$ is always a linear combination of  human-object edge features. This imposes an structure on the features processed by $\ve{R}_{E_1}$. More formally, the input into $\ve{R}_{E_1}$ is the inner product $\ve{s}^T\ve{F}$, where $\ve{F}$ is the feature matrix storing the edge features $\ve{x}_e$ s.t. $e \in E_1$. Vector $\ve{s}$ captures  the structured feature space. Its entries are in \{0,1\} depending on the node being forward-passed. In the example above $\ve{F} = [\ve{x}_{v,u}\;\; \ve{x}_{v,w}]^T$. For the human node $v$, $\ve{s}=[1\; 1]^T$, while for the object node $u$, $\ve{s}=[1\; 0]^T$.

%% file: experiment.tex
% !TEX root = main.tex
%\vspace{1\sectionReduceTop}
\section{Experiment}
\vspace{0.5\sectionReduceTop}
\label{sec:experiment}
We present results on three diverse spatio-temporal problems to ensure generic applicability of S-RNN, shown in Figure~\ref{fig:stgraphexp}. The applications include: (i) modeling human motion~\cite{Fragkiadaki15} from motion capture data~\cite{H36m}; (ii) human activity detection and anticipation~\cite{Koppula13b,Koppula13}; and (iii) maneuver anticipation from real-world driving data~\cite{Jain15}. 
%\vspace{1\subsectionReduceTop}
\subsection{Human motion modeling and forecasting}
%\vspace{1\subsectionReduceBot}
%For example, the two legs of human interact with each other to generate walking motion, while the arms collaborate to assist in eating. Furthermore, the spine connects different body parts to form a single skeleton.

Human body is a good example of separate but well related components. Its motion involves complex spatio-temporal interactions between the components (arms, legs, spine), resulting in sensible motion styles like walking, eating etc.  In this experiment, we represent the complex motion of humans over st-graphs and learn to model them with S-RNN.  We show that our structured approach outperforms the state-of-the-art unstructured deep architecture~\cite{Fragkiadaki15} on motion forecasting from motion capture (mocap) data. Several approaches based on Gaussian processes~\cite{Urtasun08,Wang08}, Restricted Boltzmann Machines (RBMs)~\cite{Taylor06,Taylor10,Sutskever09}, and RNNs~\cite{Fragkiadaki15} have been proposed to model human motion. Recently, Fragkiadaki et al.~\cite{Fragkiadaki15} proposed an encoder-RNN-decoder (ERD) which gets state-of-the-art forecasting results on H3.6m mocap data set~\cite{H36m}. 
%For example, parts of human body interact with each other to generate physically sensible motions like eating, walking etc.
\begin{figure}[t]
\centering
%\vspace{\sectionReduceTop}
\includegraphics[width=.9\linewidth]{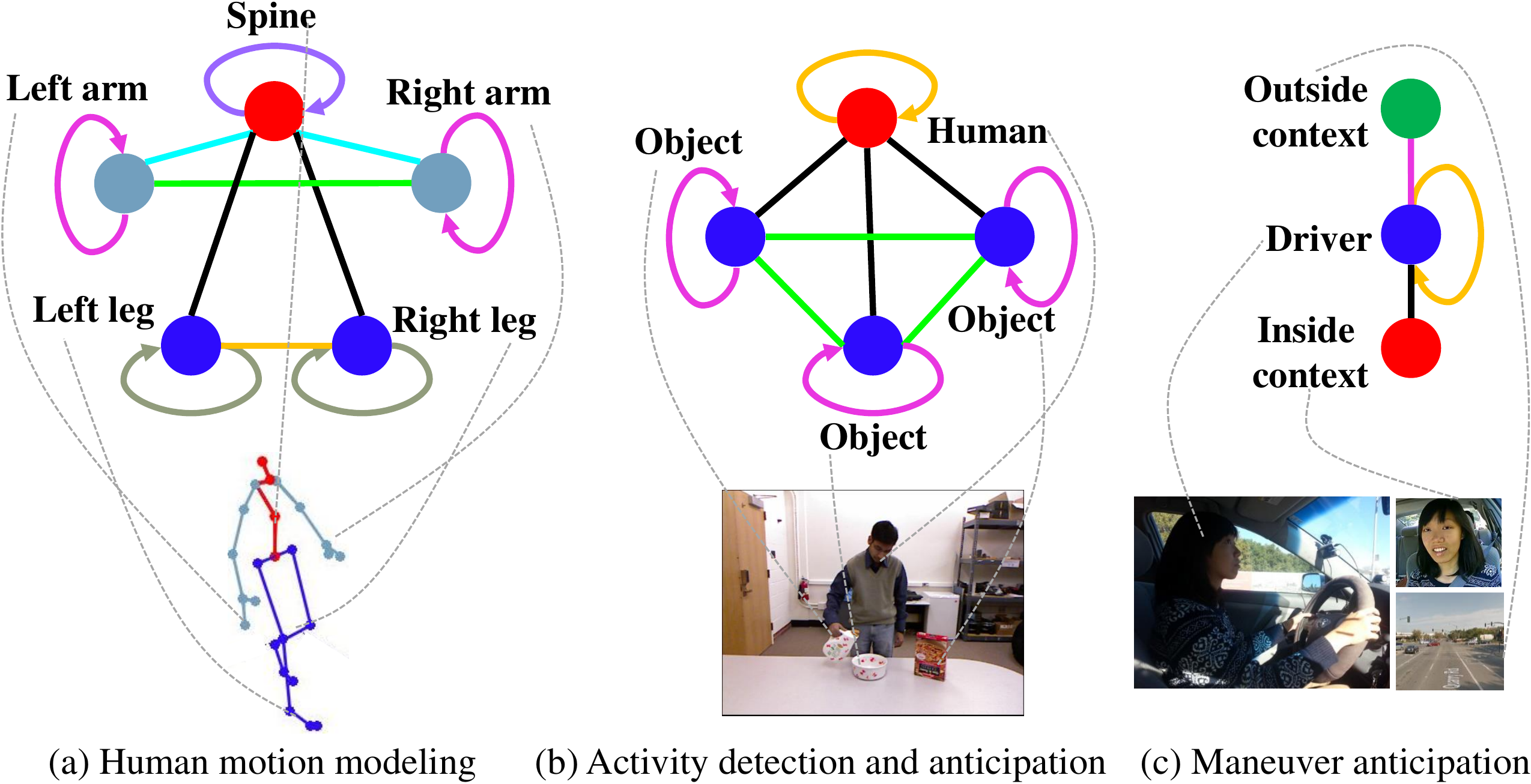}
\vspace{1.5\sectionReduceTop}
\caption{\footnotesize{\textbf{Diverse spatio-temporal tasks}. We apply S-RNN to the following three diverse spatio-temporal problems. (View in color)}}
\vspace{2\sectionReduceBot}
\label{fig:stgraphexp}
\end{figure}

\noindent \textbf{S-RNN architecture for human motion.} Our S-RNN architecture follows the st-graph shown in Figure~\ref{fig:stgraphexp}\rc{a}. According to the st-graph, the spine interacts with all the body parts, and the arms and legs interact with each other. The st-graph is automatically transformed to S-RNN following Section~\ref{subsec:srnn}. The resulting S-RNN have three nodeRNNs, one for each type of body part (spine, arm, and leg), four edgeRNNs for modeling the spatio-temporal interactions between them, and three edgeRNNs for their temporal connections. For edgeRNNs and nodeRNNs we use FC(256)-FC(256)-LSTM(512) and LSTM(512)-FC(256)-FC(100)-FC($\cdot$) architectures, respectively, with skip input and output connections~\cite{Graves13}. The outputs of nodeRNNs are skeleton joints of different body parts, which are concatenated to reconstruct the complete skeleton. In order to model human motion, we train S-RNN to predict the mocap frame at time $t+1$  given the frame at time $t$. Similar to~\cite{Fragkiadaki15}, we gradually add noise to the mocap frames during training. This simulates curriculum learning~\cite{Bengio09} and helps in keeping the forecasted motion close to the manifold of human motion. As node features we use the raw joint values expressed as exponential map~\cite{Fragkiadaki15}, and edge features are concatenation of the node features. We train all RNNs jointly to minimize the Euclidean loss between the predicted mocap frame and the ground truth. See supplementary material on the project web page~\cite{SuppSRNN} for training details.

\begin{figure}[t]
\centering
\includegraphics[width=.9\linewidth]{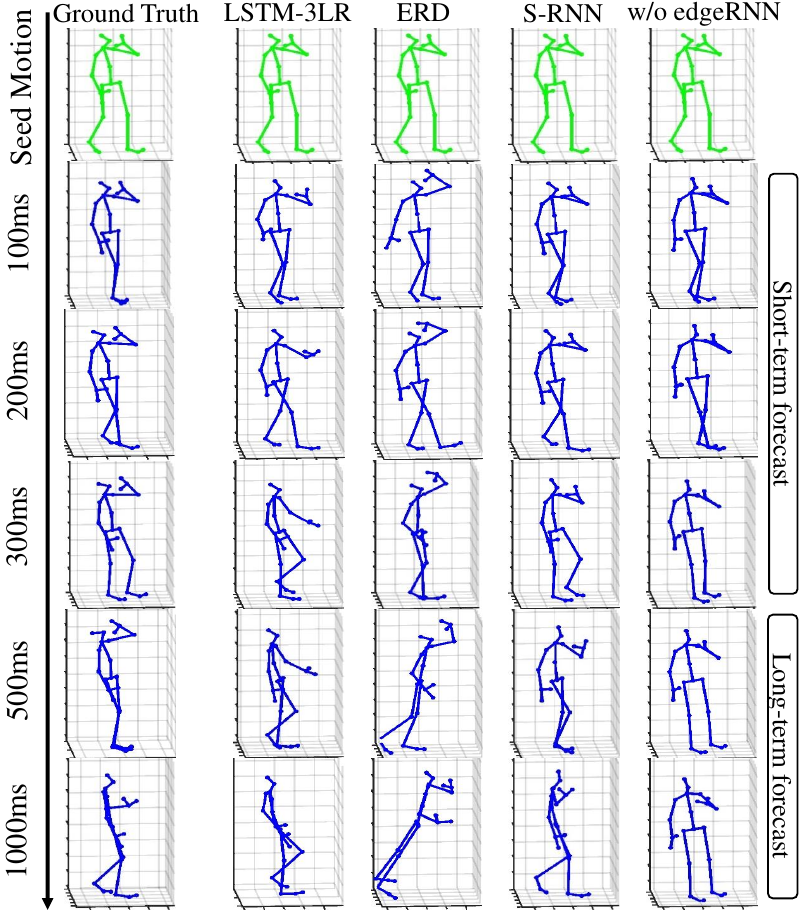}
%\vspace{1.0\sectionReduceTop}
\caption{\footnotesize{\textbf{Forecasting eating activity on test subject}. On aperiodic activities, ERD and LSTM-3LR struggle to model human motion. S-RNN, on the other hand, mimics the ground truth in the short-term and generates human like motion in the long term. Without (w/o) edgeRNNs the motion freezes to some mean standing position. See the video~\cite{SuppSRNN}.}}
\label{fig:eating_videos}
\vspace{1.5\sectionReduceBot}
\end{figure}

%We compare S-RNN with the ERD architecture~\cite{Fragkiadaki15}, which is currently the state-of-the-art in motion forecasting on H3.6m data set~\cite{H36m}.
\noindent \textbf{Evaluation setup.} We compare S-RNN with the state-of-the-art ERD architecture~\cite{Fragkiadaki15} on H3.6m mocap data set~\cite{H36m}. We also compare with a 3 layer LSTM architecture (LSTM-3LR) which~\cite{Fragkiadaki15} use as a baseline.\footnote{We reproduce ERD and LSTM-3LR architectures following~\cite{Fragkiadaki15}. The authors implementation were not available at the time of submission.} For motion forecasting we follow the experimental setup of~\cite{Fragkiadaki15}. We downsample H3.6m by two and train on 6 subjects and test on  subject S5.   To forecast, we first feed the architectures with (50) \textit{seed} mocap frames, and then forecast the future (100) frames.  Following~\cite{Fragkiadaki15}, we consider  walking, eating, and smoking activities. In addition to these three, we also consider discussion  activity. 
%four human activities: walking, eating, smoking, and discussion.\footnote{\cite{Fragkiadaki15} considered walking, eating, and smoking activities.}

Forecasting is specially challenging on activities with complex aperiodic human motion. In H3.6m data set, significant parts of eating, smoking, and discussion activities are aperiodic, while walking activity is mostly periodic. Our evaluation demonstrates the benefits of having an underlying structure in three important ways:
%goes beyond~\cite{Fragkiadaki15} in {three important ways}: 
(i) We present visualizations and  quantitative results on complex aperiodic activities (\cite{Fragkiadaki15} evaluates only on periodic walking motion); (ii) We forecast human motion for almost twice longer than state-of-the-art~\cite{Fragkiadaki15}. This is very challenging for aperiodic activities; and finally (iii) We show S-RNN interestingly learns semantic concepts, and demonstrate its modularity by generating hybrid human motion. Unstructured deep architectures like~\cite{Fragkiadaki15} does not offer such modularity.
% In Section~\ref{sec:visualize} we visualize the memory cells of S-RNN and show they learn certain semantic concepts about human motion. 
%Such long-term forecasts are challenging for aperiodic activities, where a good model should generate plausible human-like motion in the distant future

%\caption{\footnotesize{\textbf{Motion forecasting error}. Table shows 3D angle error between the forecasted motion and the ground truth, \{80, 160, 320, 560, 1000\} msecs after the seed motion. The results are averaged over 8 seed motion sequences for each activity on the test subject.}}

\begin{table}[t]
\centering
\caption{\footnotesize{\textbf{Motion forecasting angle error}. \{80, 160, 320, 560, 1000\} msecs after the seed motion. The results are averaged over 8 seed motion sequences for each activity on the test subject.}}% S-RNN outperforms both ERD~\cite{Fragkiadaki15} and LSTM-3LR majority of times.}
\vspace{0\captionReduceBot}
\resizebox{0.8\linewidth}{!}{
\centering
\begin{tabular}{r|ccccc}\hline
\multirow{2}{*}{Methods}&\multicolumn{3}{c}{Short-term forecast}&\multicolumn{2}{c}{Long-term forecast}\\
&80ms&160ms&320ms&560ms&1000ms\\\hline
&\multicolumn{5}{c}{Walking activity}\\\hline
ERD~\cite{Fragkiadaki15}&1.30&1.56&1.84&2.00&2.38\\
LSTM-3LR&1.18&1.50&1.67&\textbf{1.81}&2.20\\
S-RNN&\textbf{1.08}&\textbf{1.34}&\textbf{1.60}&1.90&\textbf{2.13}\\\hline
&\multicolumn{5}{c}{Eating activity}\\\hline
ERD~\cite{Fragkiadaki15}&1.66&1.93&2.28&2.36&\textbf{2.41}\\
LSTM-3LR&1.36&1.79&2.29&2.49&2.82\\
S-RNN&\textbf{1.35}&\textbf{1.71}&\textbf{2.12}&\textbf{2.28}&2.58\\\hline
&\multicolumn{5}{c}{Smoking activity}\\\hline
ERD~\cite{Fragkiadaki15}&2.34&2.74&3.73&3.68&3.82\\
LSTM-3LR&2.05&2.34&3.10&3.24&3.42\\
S-RNN&\textbf{1.90}&\textbf{2.30}&\textbf{2.90}&\textbf{3.21}&\textbf{3.23}\\\hline
&\multicolumn{5}{c}{Discussion activity}\\\hline
ERD~\cite{Fragkiadaki15}&2.67&2.97&3.23&3.47&2.92\\
LSTM-3LR&2.25&2.33&2.45&{2.48}&2.93\\
S-RNN&\textbf{1.67}&\textbf{2.03}&\textbf{2.20}&\textbf{2.39}&\textbf{2.43}\\\hline
\end{tabular}
}
\vspace{1.5\sectionReduceBot}
\label{tab:3derror}
\end{table}

\noindent \textbf{Qualitative results on motion forecasting.} Figure~\ref{fig:eating_videos} shows forecasting 1000ms of human motion on ``eating'' activity -- the subject drinks while walking. S-RNN stays close to the ground-truth in the short-term and generates human like motion in the long-term. On removing edgeRNNs, the parts of human body become independent and stops interacting through parameters. Hence without edgeRNNs the skeleton freezes to some mean position. LSTM-3LR suffers with a drifting problem. On many test examples it drifts to the mean position of walking human (\cite{Fragkiadaki15} made similar observations about LSTM-3LR). The motion generated by ERD~\cite{Fragkiadaki15} stays human-like in the short-term but it drifts away to non-human like motion in the long-term. This was a common outcome of ERD on complex aperiodic activities, unlike S-RNN. Furthermore, ERD produced human motion was non-smooth  on many test examples. See the video on the project web page for more examples~\cite{SuppSRNN}.

\iffalse
S-RNN models human motion well. It stays close to the ground-truth in the short-term and generates human like motion in the long-term. Figure~\ref{fig:eating_videos}  also shows the importance of edgeRNNs, without them the human skeleton freezes to some mean position. LSTM-3LR suffers with a drifting problem. On many test examples it drifts to the mean position of walking human (\cite{Fragkiadaki15} made similar observations about LSTM-3LR). The motion generated by ERD~\cite{Fragkiadaki15} stays close to the ground-truth in the short-term but it drifts away to non-human like motion in the long-term. Furthermore, ERD produced  human motion was non-smooth on many test examples. 
%generated non-smooth This was a common outcome of ERD on aperiodic activities (eating, smoking, discussion). 
\fi

\noindent \textbf{Quantitative evaluation.} 
We follow the evaluation metric of Fragkiadaki et al.~\cite{Fragkiadaki15} and present the 3D angle error between the forecasted mocap frame and the ground truth in Table~\ref{tab:3derror}.  Qualitatively, ERD models human motion better than LSTM-3LR. However, in the short-term, it does not mimic the ground-truth as well as LSTM-3LR. Fragkiadaki et al.~\cite{Fragkiadaki15} also note this trade-off between ERD and LSTM-3LR. On the other hand, S-RNN outperforms both LSTM-3LR and ERD on short-term motion forecasting on all activities. S-RNN therefore mimics  the ground truth in the short-term and generates human like motion in the long term. In this way it well handles both short and long term forecasting. Due to stochasticity in human motion, long-term forecasts ($>$ 500ms) can significantly differ from the ground truth but still depict human-like motion. For this reason, the long-term forecast numbers in Table~\ref{tab:3derror} are not a fair representative of algorithms modeling capabilities. We also observe that discussion is one of the most challenging aperiodic activity for all algorithms. 
%Therefore S-RNN gives us the best of both worlds. It mimics  the ground truth in the short-term and generates human like motion in the long term.

\noindent \textbf{User study.} We asked users to rate the motions on a Likert scale of 1 to 3. S-RNN performed best according to the user study. See supplementary material for the results.

\iffalse
%%%%%%%%%%%%%%%%%% This part is commentted out %%%%%%%%%%%%%%%%%%
%of motion forecasting is tricky due to stochasticity in human motion. For this reason most of the previous works present qualitative results. 
We follow the evaluation metric of Fragkiadaki et al.~\cite{Fragkiadaki15} and present the 3D angle error between the forecasted mocap frame and the ground truth in Table~\ref{tab:3derror}.  For short-term motion forecasting, S-RNN outperforms both LSTM-3LR and ERD on all activities. While S-RNN produces realistic human-like motion, it also mimics the ground truth in the short-term. Qualitatively ERD models human motion better than LSTM-3LR. However, in the short-term it does mimic the ground-truth as well as LSTM-3LR. Fragkiadaki et al.~\cite{Fragkiadaki15} also note this trade-off between ERD and LSTM-3LR. Therefore S-RNN gives us the best of both world. It mimics  the ground truth in the short-term and generates human like motion in the long term. Due to stochasticity in human motion, long-term forecasts ($>$ 500ms) can significantly differ from the ground truth but they can still depict human-like motion. For this reason, the long-term forecast numbers in Table~\ref{tab:3derror} are not a fair representative of algorithms modeling capabilities. We also observe that discussion is one of the most challenging aperiodic activity for all algorithms. 
%%%%%%%%%%%%%%%%%% Uptill here %%%%%%%%%%%%%%%%%%
\fi

\noindent \textbf{To summarize}, unstructured approaches like LSTM-3LR and ERD struggles to model long-term human motion on complex activities. S-RNN's good performance is attributed to its structural modeling of human motion through the underlying st-graph.  S-RNN models each body part separately with nodeRNNs and captures interactions between  them with edgeRNNs in order to produce coherent motions.
%While the overall body motion can be aperiodic, but specific parts of the body move in an (almost) periodic manner. 

\vspace{1\subsectionReduceTop}
\subsection{Going deeper into structural-RNN}
\label{sec:visualize}
\vspace{1\subsectionReduceBot}
We now present several insights into S-RNN architecture and demonstrate the modularity of the architecture which enables it to generate hybrid human motions. 
%In particular, we visualize the memory cells and show that activations of certain cells carry semantic meanings. We also show that modularity of S-RNN enables it to generate hybrid human motions which are not in the data set. Finally, we study convergence and complexity of S-RNN architecture. 

\begin{figure}[t]
\includegraphics[width=\linewidth]{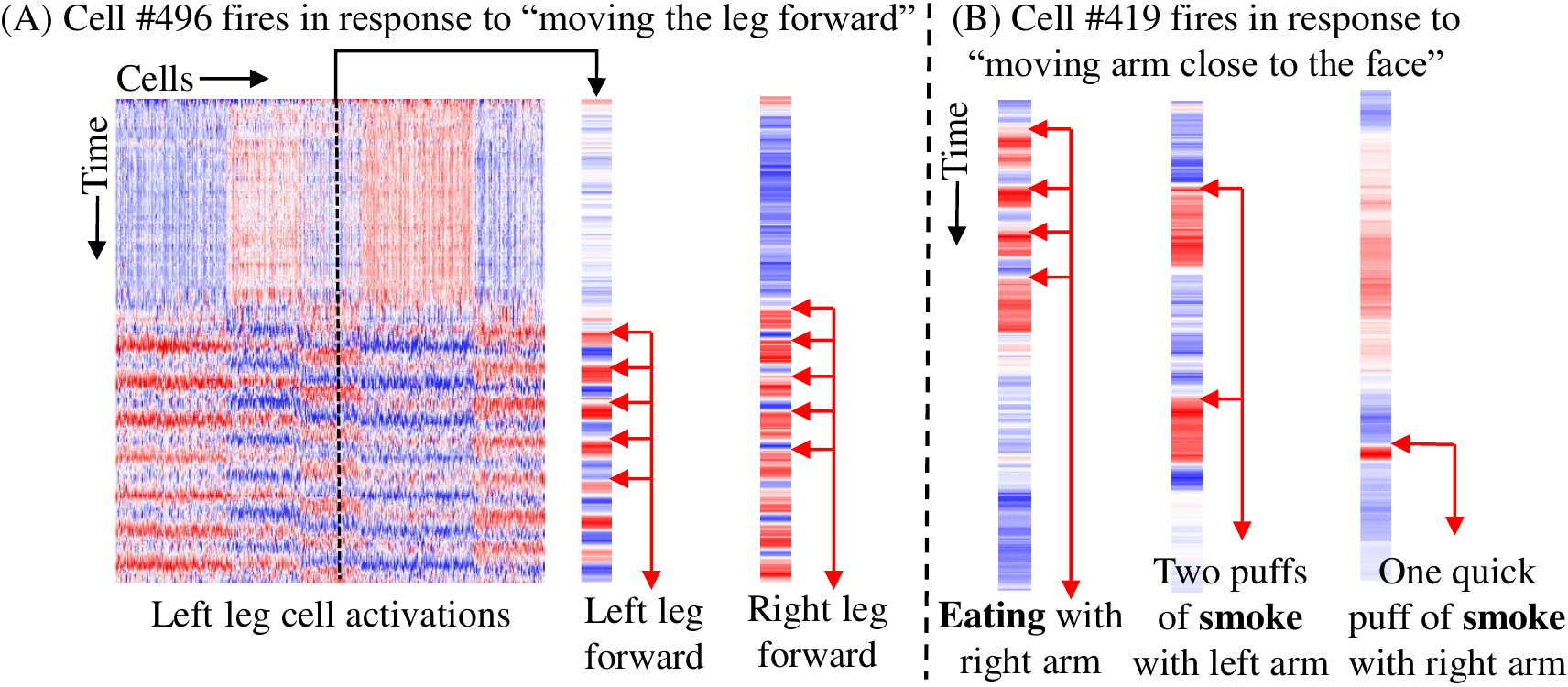}
\vspace{2.5\sectionReduceTop}
\caption{\footnotesize{\textbf{S-RNN memory cell visualization.} (\textbf{Left}) A cell of the leg nodeRNN fires (red) when ``putting the leg forward''. (\textbf{Right}) A cell of the arm nodeRNN fires for ``moving the hand close to the face''. We visualize the same cell for eating and smoking activities. (\textbf{See the video}~\cite{SuppSRNN}) }}
\vspace{2.5\sectionReduceBot}
\label{fig:memory_cell}
\end{figure}

\textbf{Visualization of memory cells.} We investigated if S-RNN memory cells represent meaningful semantic sub-motions. Semantic cells were earlier studied on other problems~\cite{Karpathy15}, we are the first to present it on a vision task and human motion. In Figure~\ref{fig:memory_cell} (left) we show a cell in the leg nodeRNN learns the semantic motion of \textit{moving the leg forward}. The cell fires positive (red color) on the forward movement of the leg and negative (blue color) on its backward movement. As the subject walks, the cell alternatively fires for the right and the left leg. Longer activations in the right leg corresponds to the longer steps taken by the subject with the right leg. Similarly, a cell in the arm nodeRNN learns the concept of \textit{moving hand close to the face}, as shown in Figure~\ref{fig:memory_cell} (right). The same cell fires whenever the subject moves the hand closer to the face during eating or smoking. The cell remains active  as long as the hand stays close to the face. See the  video~\cite{SuppSRNN}.

\iffalse
 Our architecture allows us to factor an overall complex motion into individual body parts. In Figure~\ref{fig:memory_cell} (left) we show a cell in the leg nodeRNN that \textit{learns the semantic concept of walking}. The cell fires positive (red color) on the forward movement of the leg and negative (blue color) on its backward movement. As the subject walks, the cell alternatively fires for the right and the left leg. Longer activations in the right leg corresponds to the longer steps taken by the subject with the right leg. Similarly, a cell in the arm nodeRNN \textit{learns the concept of moving hand close to the face}, as shown in Figure~\ref{fig:memory_cell} (right). The same cell fires whenever the subject moves the hand closer to the face during eating or smoking. The cell remains active  as long as the hand stays close to the face. %In Figure~\ref{fig:memory_cell} (right) we show the same cell on different activities and with varying number of movements of hand to the face.
\fi

\begin{figure}[t]
\centering
\vspace{.5\sectionReduceTop}
\includegraphics[width=0.66\linewidth]{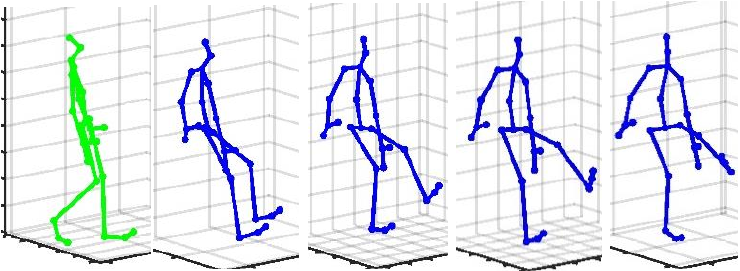}
\includegraphics[width=0.32\linewidth]{lossPlot.pdf}
\vspace{1\sectionReduceTop}
\caption{\footnotesize{\textbf{(Left)} \textbf{Generating hybrid motions} \textbf{(See the video~\cite{SuppSRNN})}. We demonstrate flexibility of S-RNN by generating a hybrid motion of a ``human jumping forward on one leg''.   \textbf{(Right)} \textbf{Train and test error}. S-RNN generalizes better than ERD with a smaller test error. }}
\vspace{1\sectionReduceBot}
%Structural nature of our architecture allows us to exchange nodeRNNs between S-RNN architectures trained on different motion styles. 
\label{fig:loss}
\end{figure}

\textbf{Generating hybrid human motion.} We now demonstrate the flexibility of our modular architecture by generating novel yet meaningful motions which are not in the data set. Such modularity is of interest and has been explored to generate diverse motion styles~\cite{Taylor09}. As a result of having an underlying high-level structure, our approach allows us to exchange RNNs between the S-RNN architectures trained on different motion styles. We leverage this to create a novel S-RNN architecture which generates a hybrid motion of a \textit{human jumping forward on one leg}, as shown in Figure~\ref{fig:loss} (Left). For this experiment we modeled the left and right leg with different nodeRNNs.  We trained two independent S-RNN models -- a slower human and a faster human (by down sampling data) -- and swapped the left leg nodeRNN of the trained models. The resulting faster human, with a slower left leg, jumps forward on the left leg to keep up with its twice faster right leg.\footnote{Imagine your motion forward if someone holds your right leg and runs!} Unstructured architectures like ERD~\cite{Fragkiadaki15} does not offer this kind of flexibility.  

Figure~\ref{fig:loss} (Right) examines the test and train error with iterations. Both S-RNN and ERD converge to similar training error, however S-RNN generalizes better with a smaller test error for next step prediction. Discussion in supplementary.

%Figure~\ref{fig:loss} (Right) examines the test and train error of S-RNN and ERD with iterations. Both methods converge to similar training error, however S-RNN converges with a smaller test error for next step prediction. Therefore S-RNN generalizes better. See supplementary for the discussion.

%\caption{\footnotesize{\textbf{(Left)} \textbf{Generating hybrid motions}. \textbf{See supplementary video}. We demonstrate flexibility of S-RNN by generating a hybrid motion of a \textit{human jumping forward on one leg}.   \textbf{(Right)} \textbf{Train and test error}. Both S-RNN and ERD have similar training error, however S-RNN generalizes better with a smaller test error for one time step prediction. }}

\begin{table*}[t]
\centering
\vspace{.75\sectionReduceTop}
\caption{\footnotesize{\textbf{Maneuver Anticipation on 1100 miles of real-world driving data}. S-RNN is derived from the st-graph shown in Figure~\ref{fig:stgraphexp}\rc{c}. Jain et al.~\cite{Jain15} use the same st-graph but models it in a probabilistic frame with AIO-HMM. The table shows average \textit{precision}, \textit{recall} and \textit{time-to-maneuver}. Time-to-maneuver is the interval between the time of algorithm's prediction and the start of the maneuver. Algorithms are compared on the features from~\cite{Jain15}. }}
\vspace{1\sectionReduceBot}
\resizebox{0.9\linewidth}{!}{
\centering
\begin{tabular}{cr|ccc|ccc|ccc}
%\hline
&  &\multicolumn{3}{c}{Turns}&\multicolumn{3}{|c}{Lane change}&\multicolumn{3}{|c}{All maneuvers}\\
\cline{1-11}
\multicolumn{2}{c|}{\multirow{2}{*}{Method}} & \multirow{2}{*}{$Pr$ (\%)}  & \multirow{2}{*}{$Re$ (\%)} & Time-to-  & \multirow{2}{*}{$Pr$ (\%)} & \multirow{2}{*}{$Re$ (\%)} & Time-to-  & \multirow{2}{*}{$Pr$ (\%)} & \multirow{2}{*}{$Re$ (\%)}  & Time-to- \\ 
& & & &  maneuver (s) &  & &  maneuver (s) &  & & maneuver (s)\\\hline
&SVM 		&	64.7 &	47.2 &	2.40 	&	73.7 &	57.8 &	2.40		&	43.7 &	37.7 & 1.20\\
&AIO-HMM	~\cite{Jain15}	 		&	{80.8}	&		75.2	&	4.16 &	{83.8}	&	79.2	&	3.80 		&	{77.4}	&	{71.2}	&	3.53 \\
 & S-RNN w/o edgeRNN  & 75.2 & 75.3  & 3.68 & 85.4  & \textbf{86.0} & 3.53 & 78.0 & 71.1 & 3.15 \\
&S-RNN  & \textbf{81.2}  & \textbf{78.6} & 3.94 & \textbf{92.7} & 84.4  & 3.46 & \textbf{82.2}  & \textbf{75.9} & 3.75 \\\hline
%\textit{Methods}&S-RNN with EL & 88.2 $\pm$ 1.4 & \textbf{86.0} $\pm$ 0.7 & 3.42 & \textbf{83.8} $\pm$ 2.1 & \textbf{79.9} $\pm$ 3.5 & 3.78 & \textbf{84.5} $\pm$ 1.0 & \textbf{77.1} $\pm$ 1.3 & 3.58\\
%\hline
\end{tabular}
}
\vspace{2\sectionReduceTop}
\label{tab:maneuver}
\end{table*}

\vspace{1\subsectionReduceTop}
\subsection{Human activity detection and anticipation}
\vspace{1\subsectionReduceBot}
In this section we present S-RNN for modeling human activities. We consider the CAD-120~\cite{Koppula13b} data set where the activities involve rich human-object interactions. Each activity consist of a sequence of sub-activities (e.g. moving, drinking etc.) and  objects affordance (e.g., reachable, drinkable etc.), which evolves as the activity progresses. Detecting and anticipating the sub-activities and affordance enables personal robots to assist humans. However, the problem is challenging as it involves complex interactions -- humans interact with multiple objects during an activity, and  objects  also interact with each other (e.g. pouring water from ``glass'' into a ``container''), which  makes it a particularly good fit for evaluating our method. Koppula et al.~\cite{Koppula13,Koppula13b} represents such rich spatio-temporal interactions with the st-graph shown in Figure~\ref{fig:stgraphexp}\rc{b}, and models it with a spatio-temporal CRF. In this experiment, we show that modeling the same st-graph with S-RNN yields  superior results. We use the node and edges features from~\cite{Koppula13b}.

Figure~\ref{fig:stgraph}\rc{b} shows our S-RNN architecture  to model the st-graph. Since the number of objects varies with environment, factor sharing between the object nodes and the human-object edges becomes crucial. In S-RNN, $\ve{R}_{V_2}$ and $\ve{R}_{E_1}$ handles all the object nodes and the human-object edges respectively. This allows our fixed S-RNN architecture to handle varying size st-graphs. For edgeRNNs we use a single layer LSTM of size 128, and for nodeRNNs we use LSTM(256)-softmax($\cdot$). At each time step, the human nodeRNN outputs the sub-activity label (10 classes), and the object nodeRNN outputs the affordance (12 classes). Having observed the st-graph upto time $t$, the goal is to \textit{detect} the sub-activity and affordance labels at the current time $t$, and also \textit{anticipate} their future labels of the time step $t+1$. For detection  we train S-RNN on the labels of the current time step. For anticipation we train the architecture to predict the labels of the next time step, given the observations upto the current time. We also train a \textit{multi-task} version of S-RNN, where we add two softmax layers to each nodeRNN and jointly train for anticipation and detection.

\iffalse
In this section we present S-RNN for modeling human activities. We consider the activities in CAD-120~\cite{Koppula13b} data set. Each activity consist of a sequence of sub-activities (e.g. moving, drinking etc.) involving human-object interactions.  Koppula et al.~\cite{Koppula13,Koppula13b} represents these interactions with the st-graph shown in Figure~\ref{fig:stgraphexp}\rc{b}. They model the st-graph as an spatio-temporal CRF. The human node in the graph is labeled with the sub-activity (10 classes) and the object nodes are labeled with the affordance (12 classes). The number of object nodes in the st-graph varies with the task, subject and environment. Figure~\ref{fig:stgraph}\rc{b} shows our S-RNN architecture  to model the st-graph. All object nodes in the st-graph are handled by the nodeRNN $\ve{R}_{V_2}$. Hence a fixed S-RNN architecture can handle varying number of objects.
\fi

%\caption{\footnotesize{\textbf{Maneuver Anticipation on 1100 miles of driving data}. S-RNN is derived from the st-graph shown in Figure~\ref{fig:stgraphexp}\rc{c}. Jain et al.~\cite{Jain15} use the same st-graph but models it in a probabilistic frame with AIO-HMM. The table shows average \textit{precision}, \textit{recall} and \textit{time-to-maneuver} are computed from 5-fold cross-validation. Algorithms are compared on the features from Jain et al.~\cite{Jain15}.} }

\begin{table}[t]
\centering
\caption{\footnotesize{\textbf{Results on CAD-120~\cite{Koppula13b}}. S-RNN architecture derived from the st-graph in Figure~\ref{fig:stgraphexp}\rc{b} outperforms Koppula et al.~\cite{Koppula13,Koppula13b} which models the same st-graph in a probabilistic framework. S-RNN in multi-task setting (joint detection and anticipation) further improves the performance. }}
\vspace{1\sectionReduceBot}
\resizebox{1\linewidth}{!}{
\centering
\begin{tabular}{r|cc|cc}
&\multicolumn{2}{c|}{Detection F1-score}& \multicolumn{2}{c}{Anticipation F1-score}\\\hline
{\multirow{2}{*}{Method}} & {\multirow{1}{*}{Sub-}} & Object & {\multirow{1}{*}{Sub-}} & Object\\
&activity (\%)&Affordance (\%)&activity (\%)&Affordance (\%)\\\hline
Koppula et al.~\cite{Koppula13,Koppula13b}&80.4&81.5&37.9&36.7\\
S-RNN w/o edgeRNN &82.2&82.1&64.8&72.4\\
S-RNN &\textbf{83.2}&88.7&62.3&80.7\\
S-RNN (multi-task)&82.4&\textbf{91.1}&\textbf{65.6}&\textbf{80.9}\\\hline
\end{tabular}
}
\vspace{2.5\sectionReduceBot}
\label{tab:cad120}
\end{table}

\begin{figure}[t]
\centering
\includegraphics[width=.99\linewidth]{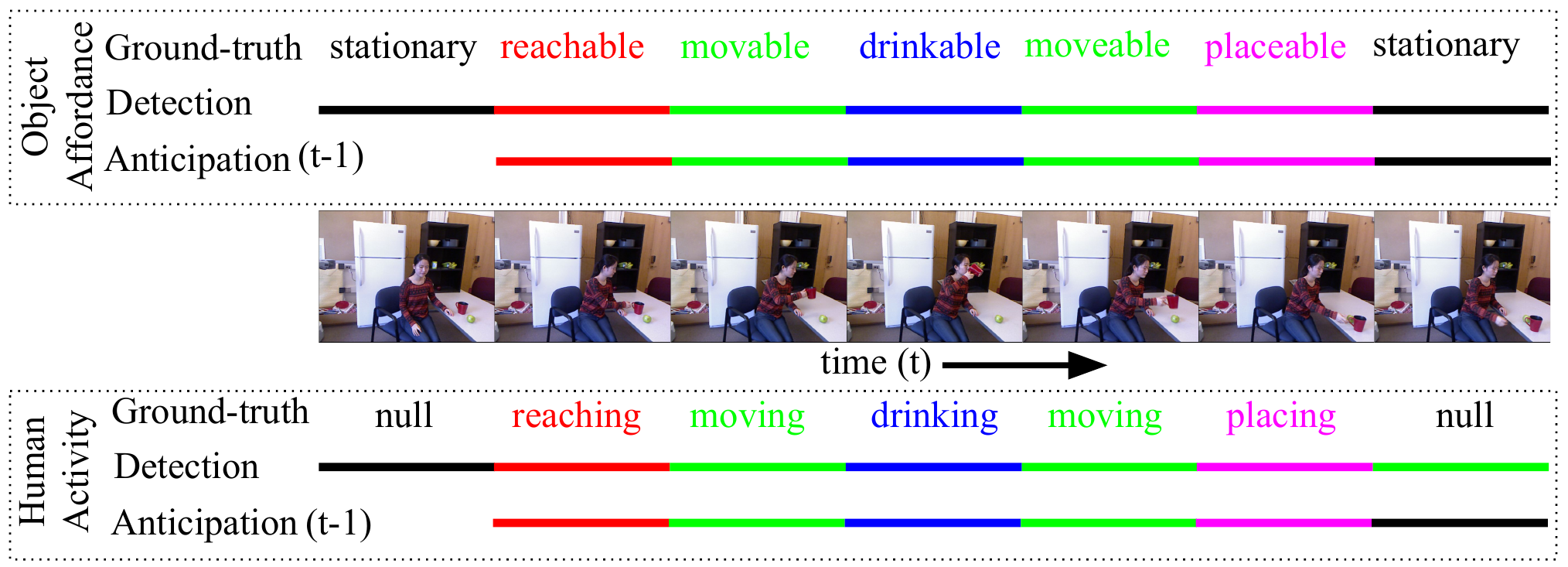}
\vspace{1.5\sectionReduceTop}
\caption{\footnotesize{\textbf{Qualitative result on eating activity on CAD-120.} Shows multi-task S-RNN detection and anticipation results. For the sub-activity at time $t$, the labels are anticipated at time $t-1$. (Zoom in to see the image) }}
\vspace{2.7\sectionReduceBot}
\label{fig:eating_anticipation}
\end{figure}

Table~\ref{tab:cad120} shows the detection and anticipation F1-scores averaged over all the classes.  S-RNN significantly improves over Koppula et al. on both anticipation~\cite{Koppula13} and detection~\cite{Koppula13b}. On anticipating object affordance S-RNN F1-score is 44\% more than~\cite{Koppula13}, and 7\% more on detection. S-RNN does not have any Markov assumptions like spatio-temporal CRF, and therefore, it better models the long-time dependencies needed for anticipation. The table also shows the importance of edgeRNNs in handling spatio-temporal components. EdgeRNN transfers the information from the human to objects, which helps is predicting the object labels. Therefore, S-RNN without the edgeRNNs poorly models the objects. This signifies the importance of edgeRNNs and also validates our design.   Finally, training S-RNN in a multi-task manner  works best in majority of the cases. In Figure~\ref{fig:eating_anticipation} we show the visualization of an eating activity. We show one representative frame from each sub-activity and our corresponding predictions. 
%Though, we observe slight improved activity anticipation without edgeRNNs.
%combines the best of the both S-RNN and S-RNN w/o edgeRNN, and

\noindent \textbf{S-RNN complexity.} In terms of complexity, we discuss two aspects as a function of the underlying st-graph: (i) the number of RNNs in the mixture; and (ii) the complexity of forward-pass. The number of RNNs depends on the number of semantically similar nodes in the st-graph. The overall S-RNN architecture is compact because the edgeRNNs are shared between the nodeRNNs, and the number of semantic categories are usually few in context-rich applications. Furthermore, because of factor sharing the number of RNNs does not increase if more semantically similar nodes are added to the st-graph. The forward-pass complexity depends on the number of RNNs. Since the forward-pass through all edgeRNNs and nodeRNNs can happen in parallel, in practice, the complexity only depends on the cascade of two neural networks (edgeRNN followed by nodeRNN).

\vspace{1\subsectionReduceTop}
\subsection{Driver maneuver anticipation}
\vspace{1\subsectionReduceBot}
We finally present S-RNN for another application which involves anticipating maneuvers several seconds before they happen.  Jain et al.~\cite{Jain15} represent this problem with the st-graph shown in Figure~\ref{fig:stgraphexp}\rc{c}. They model the st-graph as a probabilistic Bayesian network (AIO-HMM~\citep{Jain15}). The st-graph represents the interactions between the observations outside the vehicle (eg. the road features), the driver's maneuvers, and the observations inside the vehicle (eg. the driver's facial features). We model the same st-graph with S-RNN architecture using the node and edge features from Jain et al.~\cite{Jain15}. Table~\ref{tab:maneuver} shows the performance of different algorithms on this task.  S-RNN performs better than the  state-of-the-art AIO-HMM~\cite{Jain15} in every setting. See supplementary material for the discussion and details~\cite{SuppSRNN}.

%% file: conclusion.tex
% !TEX root = main.tex
\vspace{1.3\sectionReduceTop}
\section{Conclusion}
\vspace{\sectionReduceBot}

We proposed a generic and principled approach for combining high-level
spatio-temporal graphs with sequence modeling success of RNNs. We make use of
factor graph, and factor sharing in order to obtain an RNN mixture that is
scalable and applicable to any problem expressed over st-graphs. Our RNN mixture
captures the rich interactions in the underlying st-graph. We demonstrated
significant improvements with S-RNN on three diverse spatio-temporal problems
including: (i) human motion modeling; (ii) human-object interaction; and (iii)
driver maneuver anticipation. By visualizing the memory cells we showed that
S-RNN learns certain semantic sub-motions, and demonstrated its modularity by
generating novel human motion.\footnote{We acknowledge NRI \#1426452,
ONR-N00014-14-1-0156, MURI-WF911NF-15-1-0479 and Panasonic Center grant \#122282.}

\iffalse
We proposed a generic and principled approach for combining high-level spatio-temporal graphs with sequence modeling success of RNNs. We make use of factor graph, and factor sharing in order to obtain an RNN mixture that is scalable and applicable to any problem expressed over st-graphs. Our RNN mixture captures the rich interactions in the underlying st-graph through connections between the RNNs. It learns the dependencies between the output labels by sharing RNNs between the outputs. 

We demonstrated significant improvements with S-RNN on three diverse spatio-temporal problems. We showed that representing human motion over st-graph and learning via S-RNN outperforms state-of-the-art  RNN based methods. We further showed, on two context-rich spatio-temporal problems: (i) human-object interaction; and (ii) driver maneuver anticipation; that learning S-RNN from their st-graphs outperforms the existing state-of-the-art non-deep learning based methods on the same st-graph. By visualizing the memory cells we showed that S-RNN learns certain semantic sub-motions, and demonstrated its modularity by generating hybrid human motions. Future work includes combining S-RNN with CNNs for spatio-temporal feature learning~\citep{Tran15}, and developing inference methods on S-RNN for structured-output prediction.
\fi